\documentclass[twocolumn]{IEEEtran}

\usepackage{graphicx}
\usepackage{amsmath,amssymb,xspace,lscape,here,epsfig,psfrag}
\usepackage{latexsym}
\newcommand{\Poincare}{Poincar\'e}

\newcommand{\D}{{\mathcal D}}

\newcommand {\qi}{q_{i}}
\newcommand {\dotqi}{\dot{q}_{i}}
\newcommand {\qf}{q_{f}}
\newcommand {\dotqf}{\dot q_{f}}

\newcommand {\qizero}{(q_{i})_{0,st}}  

\newcommand {\qoneonei}{(q_{i})_{1}}  
\newcommand {\dotqoneonei}{(\dot {q}_{i})_{1}}  
\newcommand {\qoneonef}{(q_{f})_{1}}  
\newcommand {\dotqoneonef}{(\dot q_{f})_{1}}  

\newcommand{\thetaonei}{\theta_{i}}
\newcommand{\dotthetaonei}{\dot{\theta}_{i}}
\newcommand{\thetaonef}{\theta_{f}}
\newcommand{\dotthetaonef}{\dot{\theta}_{f}}

\newcommand{\qaonei}{(q_{i})_a}
\newcommand{\dotqaonei}{(\dot{q}_{i})_a}
\newcommand{\qaonef}{(q_{f})_a}
\newcommand{\dotqaonef}{(\dot{q}_{f})_a}

\begin{document}

\date{}

\title{ \LARGE \bf Asymptotically Stable Walking of a Five-Link Underactuated 3D Bipedal Robot}

\author{Christine Chevallereau, J.W.~Grizzle, and Ching-Long Shih
\thanks{Christine Chevallereau is with the CNRS, IRCCyN, Nantes Atlantic University, 1 rue de la Noe, 44321
Nantes, cedex 03, France,
\texttt{Christine.Chevallereau@irccyn.ec-nantes.fr}}
\thanks{\textbf{Corresponding Author:} Jessy W. Grizzle is with the Control Systems Laboratory, Electrical Engineering and Computer
      Science Department, University of Michigan, Ann Arbor, MI
      48109-2122, USA, \texttt{grizzle@umich.edu.}}
\thanks{Ching-Long Shih is with the EE Department, National Taiwan University of Science and Technology, Taipei, Taiwan
106, \texttt{shihcl@mail.ntust.edu.tw} }
}

\markboth{\bf Regular Paper Submitted to IEEE TRO: 8 January 2008; Revised 18 June 2008}{}


\maketitle

\begin{abstract} This paper presents three feedback controllers that achieve an asymptotically stable, periodic, and
fast walking gait for a 3D bipedal robot consisting of a torso,
revolute knees, and passive (unactuated) point feet. The walking
surface is assumed to be rigid and flat; the contact between the
robot and the walking surface is assumed to inhibit yaw rotation.
The studied robot has 8 DOF in the single support phase and 6 actuators.
In addition to the reduced number of actuators, the interest of
studying robots with point feet is that the feedback control
solution must explicitly account for the robot's natural dynamics in order to
achieve balance while walking. We use an extension of the method of
virtual constraints and hybrid zero dynamics, a very successful
method for planar bipeds, in order to simultaneously compute a
periodic orbit and an autonomous feedback controller that realizes
the orbit, for a 3D (spatial) bipedal walking robot. This method
allows the computations for the controller design and the periodic
orbit to be carried out on a 2-DOF subsystem of the 8-DOF robot
model. The stability of the walking gait under closed-loop control is evaluated with the  linearization of the restricted \Poincare\ map of the
hybrid zero dynamics. Most periodic walking gaits for this robot are unstable when the controlled outputs are selected to be the actuated coordinates. Three strategies are explored to produce stable walking. The first strategy consists of imposing a stability condition during the search of a periodic gait by optimization. The second strategy uses an event-based controller to modify the eigenvalues of the (linearized) \Poincare\ map. In the third approach, the effect of output selection on the zero dynamics is discussed and a pertinent choice of outputs is proposed, leading to stabilization without the use of a supplemental event-based controller.
\end{abstract}

\section{Introduction}
\label{sec:SGC:Intro} The primary objective of this paper\footnote{A preliminary version of this paper was presented in \cite{SGC07}.} is to contribute to the feedback control
of 3D bipedal robots that do not rely on large feet and slow movement for achieving stability of a
walking gait. We assume here an unactuated point contact at the leg end and, for a simple 5-link
robot, seek a time-invariant feedback controller that creates an exponentially stable, periodic
walking motion. Our approach is based on an extension of the method of virtual constraints, which
was developed in \cite{GRABPL01,PLGRWEAB03,CHABAOPLWECAGR02,WEGRKO03} for planar robots, and is extended here to the case of spatial robots.
Virtual constraints are holonomic constraints on the robot's configuration that are asymptotically
achieved through the action of a feedback controller. Their function is to coordinate the evolution
of the various links of the robot throughout a stride---which is another way of saying that they
reduce the degrees of freedom. By using virtual constraints to achieve link coordination on a
bipedal robot, different gaits can be more easily programmed than if the links were coordinated by
hardware constraints.

The work most closely related to ours is \cite{FUDOHAKA06}, where the control of a 3D walker was decomposed into
the study of its motion in the sagittal plane and the frontal plane; see also \cite{KUO99} for a related
decomposition result on control in the frontal plane. The method of virtual constraints was applied
in \cite{FUDOHAKA06} to regulate the sagittal plane motion of the biped, while an inverted pendulum approximation of
the dynamics was used to design a controller for the frontal plane. An event-based controller was
then introduced to synchronize the phasing of the independently designed sagittal and frontal
plane controllers. The overall closed-loop system was shown to be stable through simulation and
subsequently through experimentation. In our approach, we do not decompose the model into sagittal
and frontal plane motions, and coupling of the sagittal and frontal plane dynamics is introduced
into the controller from the very beginning.

A very interesting study of the feedback control of underactuated spatial robots has been given in \cite{SOZE06},  where a controller for a
five-link 3D robot with unactuated point feet has been designed on the basis of linearizing the
robot's dynamic model along a periodic orbit. So that the controller would be time-invariant, the orbit was
parameterized with a configuration variable that is strictly monotonic throughout a normal gait, as
in \cite{GRABPL01,PLGRWEAB03,CHABAOPLWECAGR02,WEGRKO03}, before linearization was applied. The (within-stride) control law is designed on the basis of a discrete-time approximation of the linearized model, which makes stability of the closed-loop system difficult to assess.

Other important work includes \cite{CORUTEWI05} and references therein, where the analysis of
passive spatial bipeds is presented. The emphasis in their work is on energy efficiency and underactuation; the role of feedback control in achieving a wide range of behaviors is not emphasized.
On the other hand, the work in \cite{SPBU05,AMGR07} seeks energy efficiency and a
large basin of attraction under the assumption of full actuation; in particular, full actuation between the
leg and ground is assumed (pitch, roll and yaw), as opposed to the unactuated
assumption made here. Very careful stability analysis of the closed-loop system is provided through geometric (Routhian) reduction. This work is taken one step further in \cite{GRSP07}, where, starting from a 2D (sagittal plane) passive limit cycle,
the authors use geometric reduction to first achieve control of the frontal plane motion and then a second stage of geometric reduction to achieve steering within the walking surface.

To the best of our knowledge, other work on the control of spatial robots either assumes full
actuation or does not provide significant analysis of the closed-loop system. There are many
control strategies based on the zero moment point ZMP \cite{VUBOPO06}, with one of the more famous
users being the robot ASIMO \cite{TaEgToTaKa88}. In this approach, a desired trajectory of the ZMP
is defined and successive inner control loops are closed on the basis of the ZMP. In the work of
\cite{KaMaHaKaKaFuHi03}, predictive control is performed on the basis of the position of the center
of mass and a simplified model of the robot in order to achieve a desired ZMP trajectory. Recently, on-line
adjustment of the ZMP has been added \cite{KaKaKaFuHaYoHi06}; this control method is implemented on
the robot HRP2. The control of the ZMP ensures that the supporting foot will not rotate about its
extremities, but this does not ensure stability in the sense of convergence toward a periodic
motion, as proved in \cite{CHOIJ05a}.

\section{Model}
\label{sec:SGC:model}

A simplified model of a spatial bipedal robot is given here. The model was chosen to be complex enough to capture interesting features of gait control that do not occur in planar robots, and simple enough that the presentation of the ideas will remain transparent. It is our expectation that the ideas presented here apply to a wider class of bipeds, but proving such a conjecture is not the objective of this paper.

\subsection{Description of the robot and the walking gait}
\begin{figure}[htbp]
\centerline{\includegraphics[width=2.in]{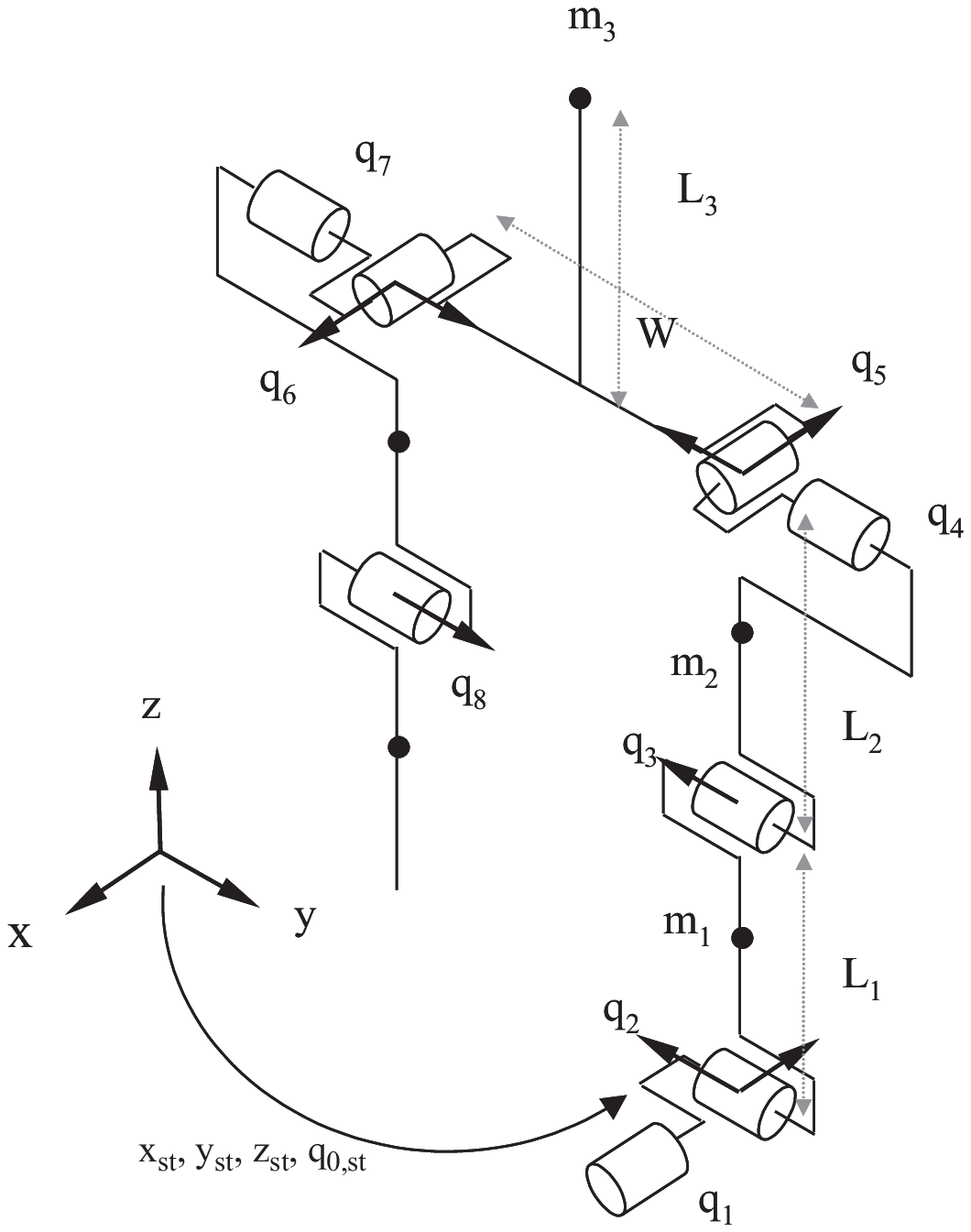}}
\caption{A five-link 3D biped with point feet in support on leg-1. There is no yaw motion about the
stance leg end and the DOF at the leg end are unactuated. For simplicity, each link is modelled by
a point mass at its center.}
\label{fig_conf1}
\end{figure}

The 3D bipedal robot discussed in this work is depicted in Fig.~\ref{fig_conf1}. It
consists of five links: a torso and two legs with revolute one DOF knees that are independently
actuated and terminated with ``point-feet''.
Each hip consists of a revolute joint with two degrees
of freedom and each degree of freedom (DOF) is independently actuated. The width of the hips is nonzero. The stance leg is assumed to act as a passive pivot in the sagittal and frontal planes, with no
rotation about the z-axis (i.e., no yaw motion), so the leg end is modeled as a point contact with
two DOF and no actuation. As discussed below, this model corresponds to the limiting case of robot with feet when the size of the feet decreases to zero. The unactuated DOF at the leg ends correspond to the classical DOF of an ankle. The DOF corresponding to the swing-leg ankle are not modelled. In total, the biped in the single support phase has eight
DOF, and there are two degrees of underactuation.

In a more complete model with feet, one would include at a minimum two degrees of freedom in the  ankles, corresponding to motion in the sagittal and frontal planes. Moreover, it would be assumed that the friction between the foot and the ground is sufficient to prevent sliding, and hence in particular, rotation of the foot in the yaw direction \cite{FUKA98,FUSA90}. In most control studies, which assume flat-footed walking, the key limiting factor is the determination of ankle torques that respect the ZMP condition, namely the ground reaction forces must remain with the convex hull of the foot \cite{GOS99,VUBOSUST90,VUBOPO06}. The lighter and smaller feet, the tighter are the constraints on the allowable ankle torques, increasing the difficulty of determining feasible walking trajectories and subsequently, stabilizing feedback controllers.

The objective of our study is two-fold: to show that stable flat-footed walking is possible with zero ankle torques, thereby removing an important obstacle to previous studies on walking; and to prepare for a future study of walking gaits that allow phases with rotation about the heel and toe \cite{ChDjGr08,CHOIJ05a}. In the limiting case of a point foot, the allowable ankle torque becomes zero, leading to an unactuated contact at the leg end. Given that the envisioned application of our results is to robots with feet, where yaw rotation is naturally inhibited by friction, it makes sense to assume in the point-foot model that yaw rotation is not allowed.

In summary, the following assumptions are made in the present study:
\begin{itemize}
\item Each link is rigid and has mass.
\item Walking consists of two alternating phases of motion: single support and double support.
\item The double support phase is instantaneous and occurs when the swing leg impacts the
ground.
\item At impact, the swing leg neither slips nor rebounds.
\item The swing and stance legs exchange their roles at each impact.
\item The gait is symmetric in steady state.
\item Walking takes place on a flat surface.
\end{itemize}
A more detailed list of hypotheses is given in \cite[Chap.~3]{WGCCM07} for planar robots, and with the obvious modifications for spatial robots, those hypotheses apply equally well here.

Since the gait is composed primarily of single support phases, the variables used to describe the robot are adapted to this phase of motion. The robot is represented as a tree structure. The stance foot, which is fixed on the ground, is the base of the tree structure. A set of generalized coordinates $q=[q_1, \ldots, q_8]'$  is shown in Fig.~\ref{fig_conf1}. Absolute angles $(q_1,
q_2)$ are roll and pitch angles of the stance leg, respectively. Angles $q_3$ and $q_8$ are the
relative joint angles of the stance-leg knee and swing-leg knee, respectively. Angles $q_4$ and
$q_5$ are the joint angles of the stance leg relative to the torso along the $y$-axis and the
$x$-axis, respectively, and angles $q_6$ and $q_7$  are the joint angles of the swing leg relative
to the torso along the $x$-axis and the $y$-axis, respectively. The coordinates $(q_1, q_2)$ are
unactuated (due to the passive contact), while $(q_3, \ldots, q_8)$ are independently actuated.

 The position of the robot with respect to an inertial frame is defined by adding the four variables $q_e =[q', x_{st}, y_{st}, z_{st}, q_{0,st} ]'$, where $x_{st}$, $y_{st}$ and $z_{st}$ are the Cartesian coordinates of the stance foot\footnote{The leg ends are referred to as feet or point feet.}, and $q_{0,st}$ defines the rotation along the z-axis of the stance leg. These variables are constant during each single support phase.

\begin{figure}[htbp]
\centerline{\includegraphics[width=2.in]{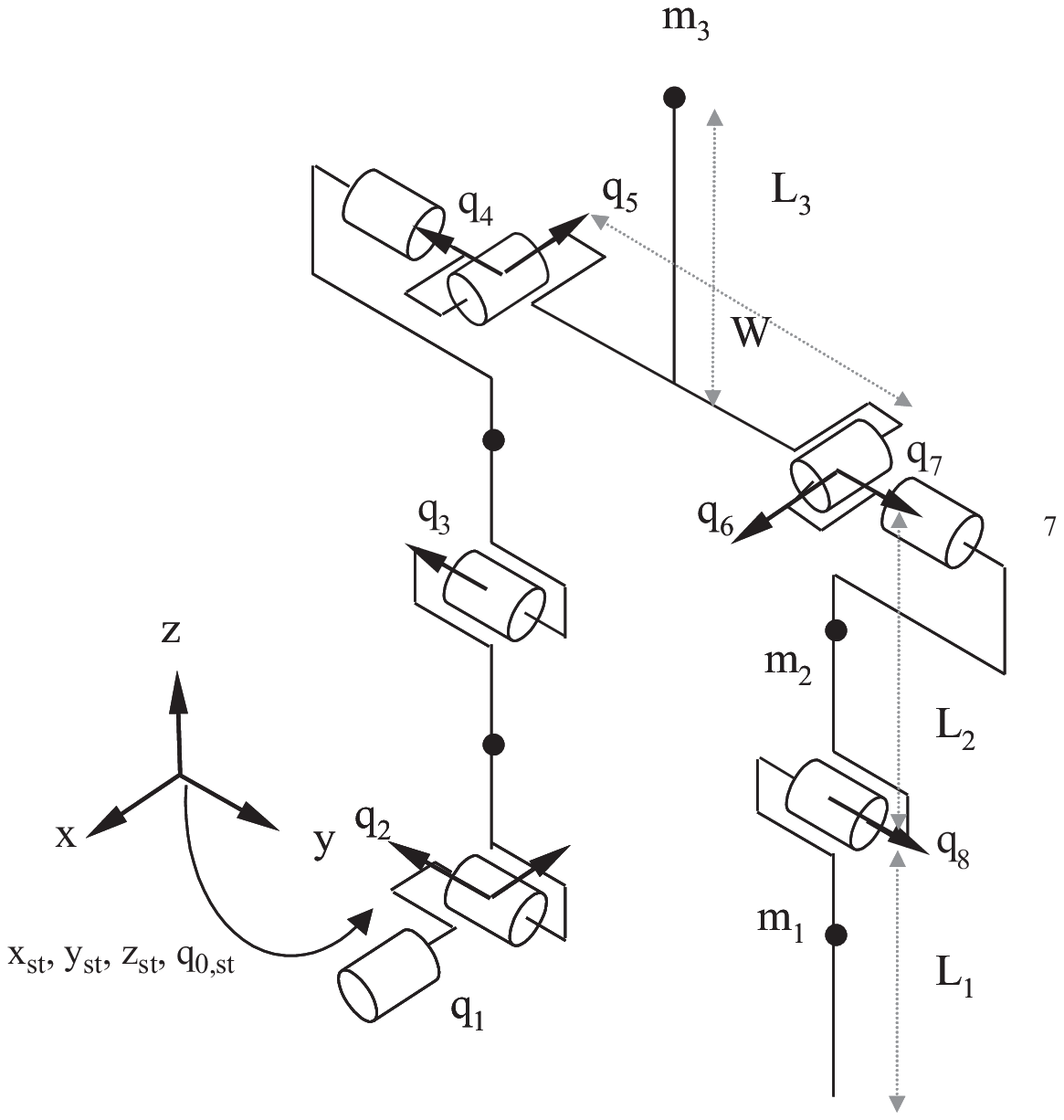}}
\caption{A five-link 3D point-feet biped in support on leg-2. The DOF at the leg end (foot) are not actuated.}
\label{fig_conf2}
\end{figure}

We have chosen to define the generalized coordinates with respect to the contact point of the current stance foot. When leg-2 is the supporting leg, the variables are defined as shown in Fig.~\ref{fig_conf2} and the same notation is employed as when the supporting leg is leg-1, viz. Fig.~\ref{fig_conf1}. Hence, at each leg exchange (i.e., impact), the variables $q_e$ undergo a jump due to the change in location of the reference frame.

During each single support phase, only one set of coordinates is used, depending on which leg is the supporting leg. In double support, either set of coordinates may be used. The transformation from one set of coordinates to the other is nonlinear \cite{SOZE06}, but it can be computed in closed form by standard means\footnote{The transformation from one set of coordinates at the end of a step, for example, to the other set of coordinates is done as follows.  Compute the orientation and the angular velocity of the swing leg shin. From this, one deduces $q_0$, $q_1$ and $q_2$ that are compatible with this orientation, and then one deduces $\dot{q}_0$, $\dot{q}_1$ and $\dot{q}_2$ yielding the angular velocity of the swing shin. The angles $q_3$ to $q_8$ exchange their roles viz $[q_3, q_4, q_5, q_6, q_7, q_8]$ $ \rightarrow$ $ [q_8, q_7, q_6, q_5, q_4, q_3]$.}.

The legs exchange roles from one step to the next. If $T$ is the duration of a step, on a periodic walking cycle, due to the choice of coordinates in Figs. \ref{fig_conf1} and \ref{fig_conf2}, we must have
\begin{equation}
\label{eq_symetrie} \begin{array}{ll}
 q_1(t+T)=-q_1(t)&q_2(t+T)=q_2(t)\\q_3(t+T)=q_3(t)&q_4(t+T)=q_4(t)\\ q_5(t+T)=-q_5(t)&q_6(t+T)=-q_6(t)\\q_7(t+T)=q_7(t)&q_8(t+T)=q_8(t)\\ \mbox{and} & \\ q_{0,st}(t+T)=-q_{0,st}(t).&\\
\end{array}
\end{equation}
The last condition yields a motion along the $x$-axis.

\begin{figure}
    \centering
    \psfrag{a}{$x$}
    \psfrag{b}{$y$}
    \includegraphics[scale=0.40]{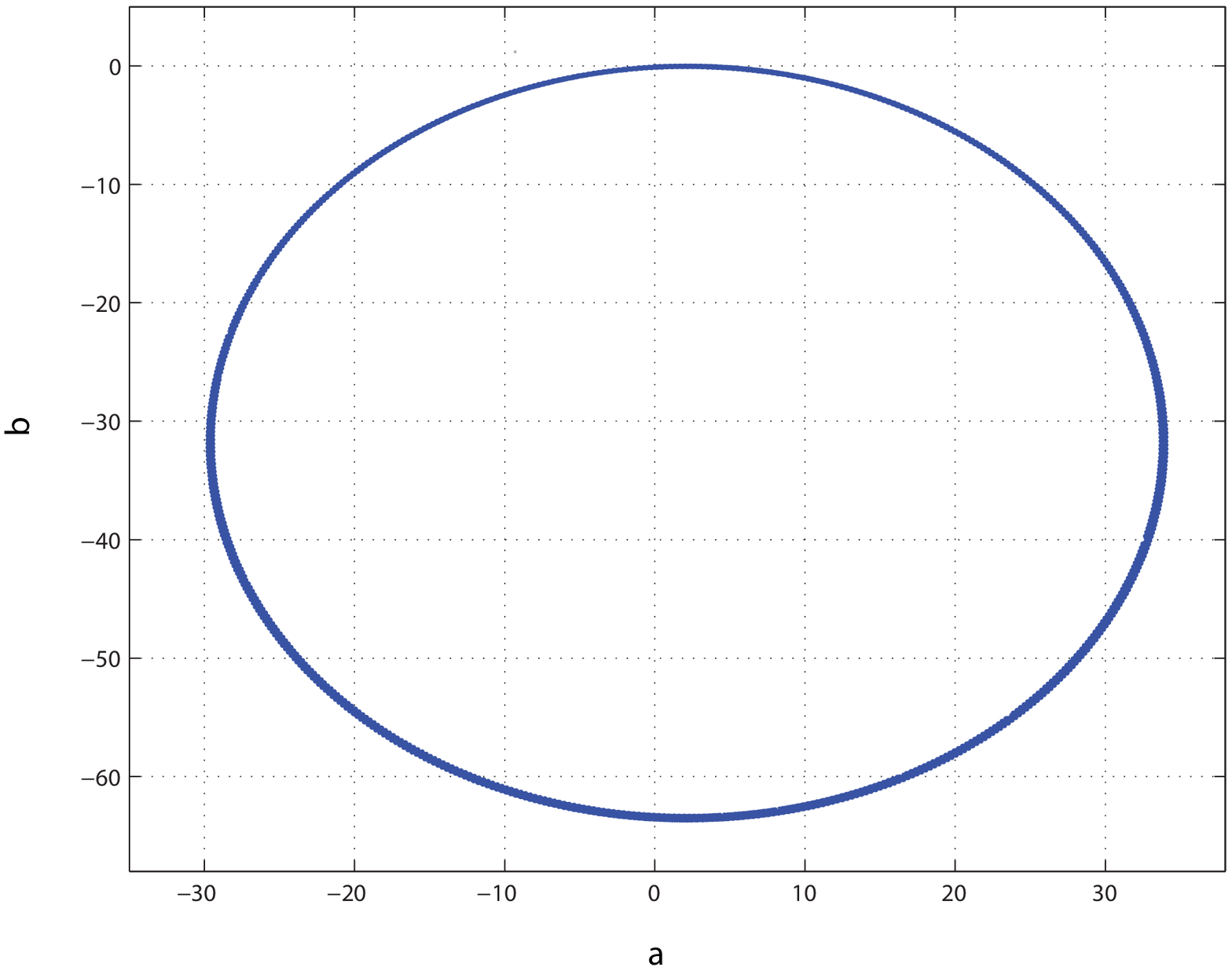}
    \caption
    {Trajectory of the robot's center of mass projected onto the $x$-$y$-plane, showing that a net rotation is possible even in a model without explicit yaw rotation. This motion arises from rotational coupling between the frontal and sagittal planes.}
    \label{fig:circle}
\end{figure}

\noindent \textbf{Remark:} Even though the model does not include rotation about the supporting foot, nor any other vertical axis, such as at the hip, coupling between the rotations in the sagittal and frontal planes can yield a net rotation about the vertical axis from one step to the next. Hence it is important to keep track of $q_{0,st}$. As an example, Fig. \ref{fig:circle} shows the projection onto the $x$-$y$-plane of a solution of the model yielding a circular motion. This was obtained by modifying the control law of \ref{sec:improved_outputs} to have unequal step lengths on the right and left legs.

\subsection{Dynamic model}
The dynamic models for single support and impact (i.e., double support) are derived here assuming support on leg $1$. The models for support on leg $2$ can be written in a similar way by using a hip width of $-W$ in place of $W$ (this places the swing leg on correct side of the stance leg).
The Euler-Lagrange equations yield the dynamic model for the robot
in the single support phase as
\begin{equation}
\label{eq_mod_dyn} D(q) \ddot {q} +H(q,\dot
{q})
=B \,u =\left[ {{\begin{array}{*{20}c}
 {0_{2\times 6} } \hfill \\
 {I_{6\times 6} } \hfill \\
\end{array} }} \right]\,u ,
\end{equation}
where $D(q)$ is the positive-definite ${(8\times 8)}$ mass-inertia matrix, $H(q,\dot {q})$ is the ${(8\times 1)}$ vector of Coriolis and gravity terms, $B$ is an ${(8\times 6)}$ full-rank, constant matrix indicating whether a joint is actuated or not,
and $u$ is the ${(6\times 1)}$ vector of input torques.
Following standard practice in the literature, the double support
phase is assumed to be instantaneous. However, it actually consists
of two distinct subphases: the impact, during which a rigid impact
takes place between the swing foot and the ground, and coordinate
relabeling. During the impact, the biped's
configuration variables do not change, but the generalized
velocities undergo a jump. The derivation of the impact model
in double support phase requires the use of the vector
$q_e$. Conservation of angular momentum of the robot about the end of the swing leg
during the impact process, in combination with the swing leg neither slipping nor
rebounding at impact, yields
\begin{equation}
\label{eq_impact} \left[ {{\begin{array}{*{20}c}
 {\dot {q}_e^+ } \hfill \\
 {F_{sw} } \hfill \\
\end{array} }} \right]=\left[ {{\begin{array}{*{20}c}
 {D_e } \hfill & {-E_{sw}' } \hfill \\
 {E_{sw} } \hfill & {0_{4\times 4} } \hfill \\
\end{array} }} \right]^{-1}\,\left[ {{\begin{array}{*{20}c}
 {D_e \dot {q}_e^- } \hfill \\
 {0_{4\times 1} } \hfill \\
\end{array} }} \right],
\end{equation}
where $\dot {q}_e^- $ and $\dot {q}_e^+ $ are the extended
velocities before and after the impact, respectively, $F_{sw}$ is the
reaction force at the contact point, $D_e$ is the extend
mass-inertia matrix, and $E_{sw} =\frac{\partial }{\partial q_e }\left[
{x_{sw}, y_{sw}, z_{sw}, q_{0,sw} } \right]'$ is the Jacobian matrix for
the position of the swing foot and its orientation in the $x-y$-plane.
Analogously to \cite{GRABPL01}, the overall impact model is written as
\begin{equation}
\label{eq_imp_pos} q^+=\Delta _q (q^-)
\end{equation}
and
\begin{equation}
\label{eq_imp_vit} \dot {q}^+=\Delta _{\dot {q}} (q^-,\dot {q}^-),
\end{equation}
and is obtained from solving (\ref{eq_impact}) and projecting down to the
generalized coordinates for support on leg $2$.

\begin{figure}
    \centering
    \psfrag{a}{\footnotesize$z_{sw}(q)=0\; \& \;x_{sw}(q)>0$}
    \psfrag{b}{\scriptsize$\dot x_1 = f_1(x_1)+g_1(x_1)u_1$}
    \psfrag{c}{\footnotesize$x_2^+=\Delta_1(x_1^-)$}
    \psfrag{d}{\footnotesize$z_{sw}(q)=0\; \& \; x_{sw}(q)>0$}
    \psfrag{e}{\scriptsize$\dot x_2 = f_2(x_2)+g_2(x_2)u_2$}
    \psfrag{f}{\footnotesize$x_1^+=\Delta_2(x_2^-)$}
    \includegraphics[scale=0.62]{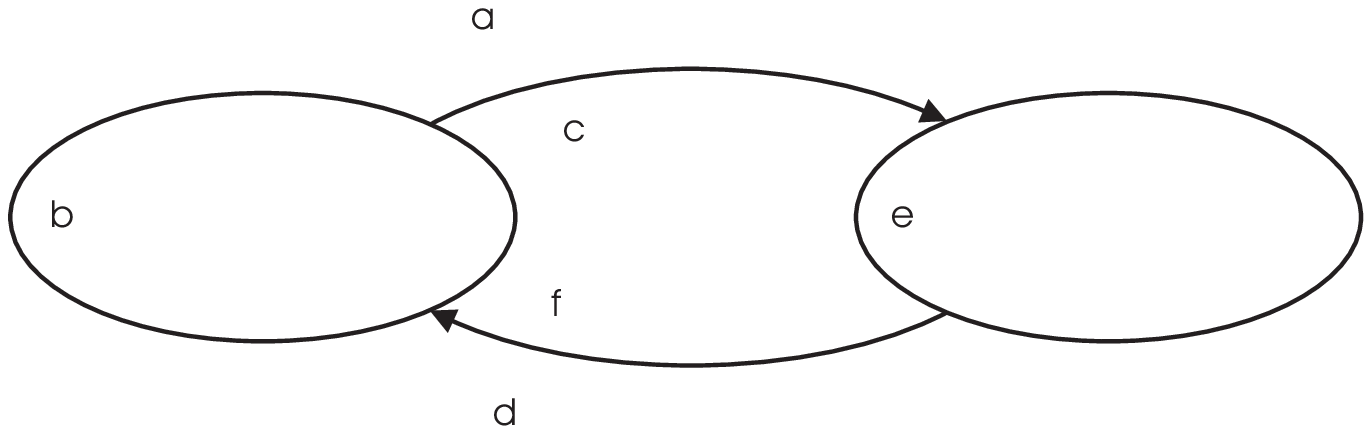}
    \caption
    {Bipedal robot's dynamic model as a hybrid system.}
    \label{hybridtwomode}
  \end{figure}

Define state variables as $x_j=\left[ {{\begin{array}{*{20}c}
 q \hfill \\
 {\dot {q}} \hfill \\
\end{array} }} \right]$, and let $x_j^+=\left[ {{\begin{array}{*{20}c}
 {q^+} \hfill \\
 {\dot {q}^+} \hfill \\
\end{array} }} \right]$and $x_j^-=\left[ {{\begin{array}{*{20}c}
 {q^-} \hfill \\
 {\dot {q}^-} \hfill \\
\end{array} }} \right]$, where the subscript $j \in \{1,2\}$ denotes the stance leg number. Then a complete walking motion of the robot can be
expressed as a nonlinear system with impulse effects, as shown in
Fig.~\ref{hybridtwomode} and written as
\begin{equation}
\label{eq_mod_hybride} \Sigma :\left\{ {{\begin{array}{*{20}c}
 {\dot x_{1}=f_1(x_1)+g_1(x_1)\,u_1\quad x_1^-\notin S_1} \hfill \\
 {x_2^+=\Delta_1 (x_1^-)\quad \quad \quad \;x_1^-\in S_1} \hfill \\
 {\dot x_{2}=f_2(x_2)+g_2(x_2)\,u_2\quad x_2^-\notin S_2} \hfill \\
 {x_1^+=\Delta_2 (x_2^-)\quad \quad \quad \;x_2^-\in S_2} \hfill \\
\end{array} }} \right.,
\end{equation}
where $S_1=\{(q,\dot {q})\vert z_{sw} (q)=0,\;x_{sw} (q)>0\}$ is the
switching surface,
\[
f_1(x)=\left[ {{\begin{array}{*{20}c}
 {\dot {q}}  \\
 {-D^{-1}(q)H(q,\dot {q})} \hfill \\
\end{array} }} \right],\quad g_1(x)=\left[ {{\begin{array}{*{20}c}
 0 \\
 {D^{-1}(q)\,B} \hfill \\
\end{array} }} \right],
\]
and
\[
x_2^+=\Delta_1 (x_1^-)=\left[ {{\begin{array}{*{20}c}
 {\Delta _q (q^-)} \hfill \\
 {\Delta _{\dot {q}} ({q}^-, \dot {q}^-)} \hfill \\
\end{array} }} \right].
\]
When leg-2 is the support leg, the same derivation produces $S_2$, $f_2$, $g_2$ and $\Delta_2$.

\section{Virtual constraints}
\label{sec_virtual_constraint}
The method of virtual constraints, which has proven very successful
in designing feedback controllers for stable walking in planar
bipeds \cite{GRABPL01,PLGRWEAB03,CHABAOPLWECAGR02,WEGRKO03}, will be applied to the 3D biped of the previous
section. In this method, one holonomic constraint per actuator is
proposed in the form of an output that, when zeroed by a feedback
controller, enforces the constraint. The most direct form of the
constraint is
\begin{equation}
\label{eq_constraint} y =h(q)=q_a -h_d (\theta ),
\end{equation}
where $q_a =[q_3 ,q_4 ,q_5 ,q_6, q_7 ,q_8]'$ is the vector of actuated
coordinates, $\theta =\theta (q)$ is a quantity that is strictly
monotonic (i.e., strictly increasing or decreasing) along a typical
walking gait, and $h_d (\theta )$ is the desired
evolution of the actuated variables as a function of $\theta $.
Roughly speaking, $\theta $ is used to replace time in
parameterizing a periodic motion of the biped. 
In a forward walking motion, the $x$-coordinate of the hip is monotonically increasing. Hence, if the virtual stance leg is defined by the line that connects the stance foot to the stance hip, the the angle of this leg in the sagittal plane is monotonic. When the shin and the thigh have the same length, the angle of the virtual leg in the sagittal plane can be selected as
\begin{equation}
\label{eq_theta_definition}
\theta =-q_2 -q_3/2
\end{equation}
(the minus sign is used to make $\theta $ strictly increasing
over a step).

The torque  $u^\ast $ required to remain on the virtual constraint surface corresponding to $q_a =h_d (\theta )$  can be computed as\footnote{As shown in \cite{SPONGM96}, \cite[pp.~60]{WGCCM07}, an expression without inversion of the $(8 \times 8)$ mass-inertia matrix $D$ is also possible. The expression given in \eqref{eq_torque_star} is more compact.}
\begin{equation}
\label{eq_torque_star} u^\ast =(\frac{\partial h(q)}{\partial
q}D^{-1}B)^{-1}\left( {\frac{\partial ^2h_d (\theta )}{\partial
\theta ^2}\dot {\theta }^2(t)+\frac{\partial h(q)}{\partial
q}D^{-1}H(q,\dot {q})} \right)
\end{equation}
This leads to an input-output linearizing controller to asymptotically drive the state of the robot to the constraint surface, assuming it does not initially start there \cite{MOGR05} \cite[Chap.~5]{WGCCM07},
\begin{equation}
\label{eq_torque} u=u^\ast -\left( {\frac{\partial h}{\partial q}D^{-1}B}
\right)^{-1}(\frac{K_p }{\varepsilon ^2}y+\frac{K_d }{\varepsilon
}\dot {y}),
\end{equation}
which results in
\begin{equation}
\label{eq_closed_loop}
    \ddot {y}+\frac{K_d }{\varepsilon }\dot
{y}+\frac{K_p }{\varepsilon ^2}y=0.
\end{equation}
In other words, determining the
constraints is equivalent to the design of a feedback controller in
the single support phase, up to the choice of the gains $K_p>0$, $K_d>0$, and $\epsilon>0$ such that \eqref{eq_closed_loop} is exponentially stable and converges sufficiently rapidly with respect to the duration of a single support phase; see \cite[Chap.~4]{WGCCM07}.

The next objective is to determine the behavior of the robot under the virtual constraints. This task is
simplified by noting that enforcing the virtual constraints,
$y =h(q)=0$, results in $q_a =h_d (\theta )$ and reduces
the dimension of the dynamics.

Let $q_u =[q_1 ,\theta ]'$
denote the unactuated joints and $q_a =[q_3 ,q_4 ,q_5 ,q_6, q_7, q_8 ]'$
denote the controlled joints, which are selected here to be the actuated joints. A linear relation exist between $q$, $q_u$ and $q_a$,
\begin{equation}
\label{eq_transf}
q={\cal T} \left[ {\begin{array}{c}
 {q_u } \\
 {q_a } \\
\end{array} } \right],
\end{equation}
where ${\cal T}$ is an $(8 \times 8)$ invertible matrix.
Then \eqref{eq_mod_dyn} can be rewritten as
\begin{equation}
\label{eq_mod_dyn_T} {\cal T}' D(q) {\cal T} \left[ {{\begin{array}{c}
 {\ddot {q}_u } \\
 {\ddot {q}_a } \\
\end{array} }} \right] + {\cal T}' H(q, \dot
q)
={\cal T}' B \,u =\left[ {{\begin{array}{*{20}c}
 {0_{2\times 6} } \hfill \\
 {I_{6\times 6} } \hfill \\
\end{array} }} \right]\,u ,
\end{equation}
The first two lines of the RHS of this equation are zero, yielding
\begin{equation}
\label{eq_mod_dyn_un}
D_{11}(q) \ddot {q}_u + D_{12}(q)\ddot {q}_a+H_1(q, \dot q)=0_{2\times 1},
\end{equation}
where $D_{11}$ is the $(2 \times 2)$ upper left sub-matrix of ${\cal T}' D(q) {\cal T}$, $D_{12}$ is the $(2 \times 6)$ upper right sub-matrix of ${\cal T}' D(q) {\cal T}$ and $H_1(q, \dot q)$ consists of the first two lines of ${\cal T}' H(q, \dot q)$.
Substituting the expressions of $q_a$, $\dot q_a$ and $\ddot q_a$ corresponding to the virtual constraints, the dynamic model of the
single support phase is now reduced to a low-dimensional, 2-DOF, autonomous
system,
\begin{equation}
\begin{array}{c}
\label{eq_dyn_zero}  D_{11}(q_u)\left[ {{\begin{array}{c}
 {\ddot q_1 } \\
 {\ddot \theta } \\
\end{array} }} \right] +D_{12}(q_u)\left({ \frac{\partial \,h_d }{\partial
\,\theta } \ddot \theta + {\frac{\partial ^2 h_d }{\partial \,\theta^2 }} \dot \theta^2} \right) \\+H_1(q_u, \dot q_u) =0,
\end{array}
\end{equation}
which is called the swing phase zero dynamics \cite{ISIDORIA95}, \cite[Chap.~5]{WGCCM07}.

One can clearly
see that the dynamic properties of the swing phase zero dynamics
depend on the particular choice of the virtual constraint $y=q_a -h_d (\theta
)=0$. How to determine a choice for $h_d (\theta )$ that results in a
periodic walking motion is summarized in the next section.

\section{Design for a Symmetric Periodic Gait}
\label{sec:optimization}

The objective of this section is to design virtual constraints $q_a =h_d(\theta )$ that correspond to a periodic motion of the robot. The gait considered is composed of single support phases separated by impacts as described in Fig.~\ref{hybridtwomode}. The legs exchange roles from one step to next, and due to symmetry, the study of a gait can be limited to a single step and the use of the symmetry relation \eqref{eq_symetrie}.

\subsection{Virtual constraints and Bezier polynomials}
The problem of designing the virtual constraints will be transformed into a parameter optimization problem as in \cite[Chap.~6]{WGCCM07}. Here, our main goal is to obtain a periodic motion; optimality is not so crucial. To simplify the optimization process, the number of variables used in the optimization problem is first reduced. This is accomplished by exploiting boundary conditions that arise from periodicity. Bezier polynomials are parametric functions that allow one to easily take into account boundary conditions on the configuration and velocity at the beginning and end of a step.

The initial and final configuration and velocity of the robot for a single support phase are important for defining the passage between the single and double support phases. Because the terminal configuration of the robot is chosen to be the instant before the double support configuration, both legs are in contact with the ground and therefore only seven independent variables are needed to describe this configuration (a closed kinematic chain). These variables parameterize the final configuration of the first step denoted $\qf$. The eight joint velocities $\dotqf$ are independent and are also added.

Knowing the final state of the single support phase, the impact model (\ref{eq_imp_pos}) and (\ref{eq_imp_vit}) determines the initial state of the ensuing single support phase. The symmetry condition \eqref{eq_symetrie} then gives the initial state of the first step: $\qi$, $\dotqi$.
The initial orientation $\qizero$ of the robot is calculated such that the  orientation for the second step is symmetric to the orientation for the first step in order that no yaw rotation is observed during the nominal (periodic) gait.

To obtain a periodic gait, the single support must be such that the state of the robot evolves from $\qi$, $\dotqi$ to $\qf$, $\dotqf$.
For given desired initial and final state values, virtual constraints can be easily deduced to connect the desired values of the actuated variables. However, the evolution of the unactuated variables is known only by integration of the dynamics \eqref{eq_dyn_zero}; a desirable dynamic behavior is imposed on these variables by the use of equality and inequality constraints in the optimization process.

\subsection{Specifics}

Here, Bezier polynomials of degree $3$ are chosen to define the virtual constraints\footnote{A degree greater than $3$ can also be chosen, in which case the number of optimization variables increases \cite{WEGRKO02}.}. The virtual constraints are expressed as functions of the variable $\theta$; see \eqref{eq_theta_definition}. From $\qi$ and $\qf$, the initial and final values of $\theta$, denoted $\thetaonei$ and $\thetaonef$,  can be calculated.
Let
\begin{equation}
\label{eq_bezier} h_d (\theta )=\sum\limits_{k=0}^3 {\alpha_k}
\frac{3!}{k!(3-k)!}s^k(1-s)^{3-k},
\end{equation}
where
\begin{equation} s=\frac{\theta -\thetaonei}{\thetaonef-\thetaonei}
\end{equation}
is the normalized independent variable. The coefficients of the  Bezier polynomials,
$\alpha_k$, are $(6 \times 1)$ vectors of real numbers. They must be determined so as to join $\qaonei$ to $\qaonef$ and $\dotqaonei$ to $\dotqaonef$, (the additional subscript ``$a$'' denotes the actuated variables) when $\theta$ varies from $\thetaonei$ to $\thetaonef$, yielding
\begin{equation}
\label{eq_coeff_Bezier}
\begin{array}{l}
\alpha_0 =h_d (\thetaonei)=\qaonei\\ \thinspace
\alpha_1 =\qaonei +\frac{\thetaonef-\thetaonei}3\frac{\partial h_d }{\partial \theta }(\thetaonei)=\qaonei
+\frac{\thetaonef-\thetaonei}3\frac{\dotqaonei }{\dot \thetaonei}\\ \thinspace
\alpha_2 =\qaonef -\frac{\thetaonef-\thetaonei}3\frac{\partial h_d }{\partial \theta }(\thetaonef)=\qaonef
-\frac{\thetaonef-\thetaonei}3\frac{\dotqaonef}{\dotthetaonef}\\ \thinspace
\alpha_3 =h_d (\thetaonef)=\qaonef. \\
 \end{array}
\end{equation}
The evolution of the unactuated variables is calculated by integration of the dynamic subsystem \eqref{eq_dyn_zero}, that is, the stance phase zero dynamics, starting from the initial state $(q_i)_u=[ \qoneonei, \thetaonei]'$ and terminating at $\theta=\thetaonef$, where $\qoneonei$ denotes the initial value of $q_1$ for the first step.

When the evolution of the unactuated variables is calculated, because the evolution of the actuated variable is given by \eqref{eq_constraint} and \eqref{eq_bezier}, the torque required to zero the constraints, i.e, $u^*$ in  \eqref{eq_torque_star}, can be calculated by lines 3 through 8 of \eqref{eq_mod_dyn_T}, and the ground reaction force $F_{st}$ expressed in the inertial reference frame (see Fig. \ref{fig_conf1}) can be calculated as well.

The search for a periodic walking motion can now be cast as a
constrained nonlinear optimization problem: Find the 15 optimization parameters prescribing $(\qf,\dotqf)$  that minimize the
integral-squared torque per step length\footnote{Torque being proportional to current in a DC motor, integral-squared torque is a rough approximation of energy dissipated in the motors.},
\begin{equation}
\label{opt_criterion}
J=\frac{1}{L}\int_{\,0}^{\,T}
{u^{\ast'} u^\ast} dt,
\end{equation}
where $T$ is the walking period
and $L$ is the step length, while satisfying symmetry \eqref{eq_symetrie}, and subject to the following:\\
\noindent \textbf{inequality constraints}
\begin{itemize}
\item $\theta $ is strictly increasing (i.e, $\dot {\theta }>0$ along
the solution);
\item  the swing foot is positioned above the ground ($z_{sw} \ge 0)$;
\item  a no-take-off constraint, $F_{st}(3) > 0 $;\smallskip
\item  a friction constraint, $\sqrt {F_{st}(1)^2 +F_{st}(2)^2 } \le \mu
\,F_{st}(3) $;
\end{itemize}

\noindent \textbf{equality constraints}\\
and a set of conditions imposing periodicity,
\begin{eqnarray*}
            q_1(T)&=&(q_f)_1 \\
             \dot q_1(T) &=& (\dot q_f)_1 \\
             \dot \theta(T) &=&\dot \theta_f,
\end{eqnarray*}
where $q_1(t)$ and $\theta(t)$ result from the integration of the zero dynamics and the walking period $T$ is such that $\theta(T)=\theta_f$.

The above procedure can be performed in MATLAB with the
\texttt{FMINCON} function of the optimization toolbox. A fixed-point solution $x^\ast=[\qf^*, \dotqf^*]'$
minimizing $J$ defines a desired periodic walking cycle (or nominal
orbit). The criterion being optimized \eqref{opt_criterion} has many local minima and the optimization technique used is local. Thus, the obtained optimal periodic motion depends on the initial set of optimization parameters.

\subsection{An example periodic motion minimizing integral-squared torque}
\label{sec_opt_motion_torque}

The physical parameters of the 3D biped studied here are given in Table \ref{tab_one}.
For these parameters, a periodic orbit was computed following
the technique presented in the previous subsection. We obtained a periodic motion defined by $x^\ast =(\qf^{\ast},\dotqf^\ast$), where
\[\begin{array}{c}\qf^\ast=[
-0.0174, -0.34038, 0.3820, -0.2940 , 0.0602, 0.0487,\\ -0.5077, 0.1688
]',
\\
\dotqf^\ast=[
 -0.4759, -1.1825, 0.0997, 0.2785, -0.1000, 0.1000,\\1.398, 0]'.
\end{array}
\]

\begin{table}[htbp]
\begin{center}
\begin{tabular}{|l|l|l|l|l|l|l|l|}
\hline g& W& L1& L2& L3& m1& m2&
m3 \\
\hline 9.81& 0.15& 0.275&0.275& 0.05& 0.875& 0.875& 5.5\\
\hline
\end{tabular}
\caption{Parameters for the 3D bipedal robot (in MKS).}
\label{tab_one}
\end{center}
\end{table}

\begin{figure}[htbp]
\centerline{\includegraphics[width=3.in]{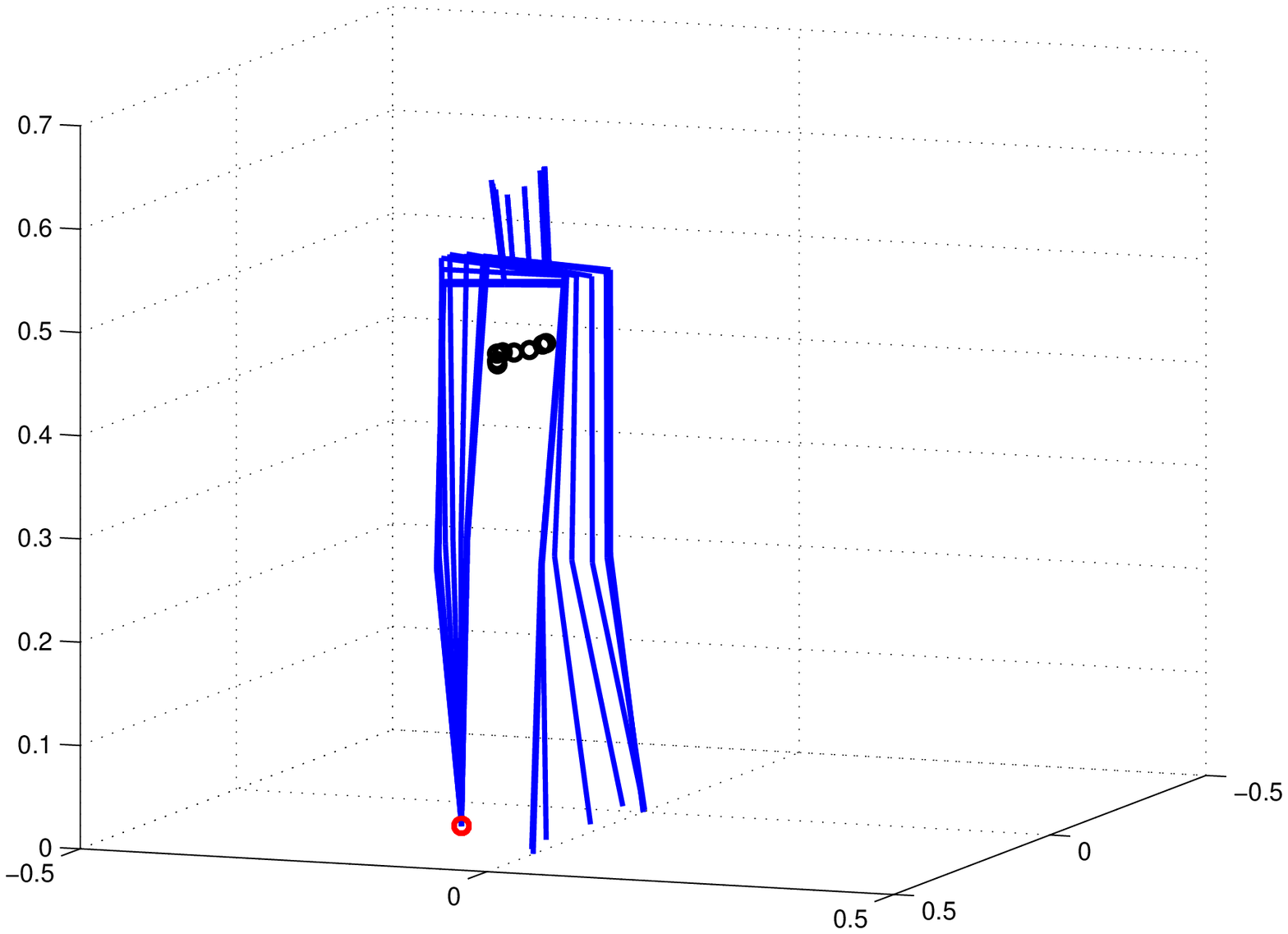}}
\caption{Stick-diagram of the optimal motion with respect to the torque criterion.}
\label{fig_stick_opt_torque}
\end{figure}

\begin{figure}[htbp]
\psfrag{q1 (deg)}{\footnotesize{$q_1$ (deg)}}
\psfrag{theta (deg)}{\footnotesize{$\theta$  (deg)}}
\psfrag{q3 (deg)}{\footnotesize{$q_3$ (deg)}}
\psfrag{q4 (deg)}{\footnotesize{$q_4$ (deg)}}
\psfrag{q5 (deg)}{\footnotesize{$q_5$ (deg)}}
\psfrag{q6 (deg)}{\footnotesize{$q_6$ (deg)}}
\psfrag{q7 (deg)}{\footnotesize{$q_7$ (deg)}}
\psfrag{q8 (deg)}{\footnotesize{$q_8$ (deg)}}
\psfrag{time (sec.)}{\footnotesize{time (s)}}
\centerline{\includegraphics[width=3.in]{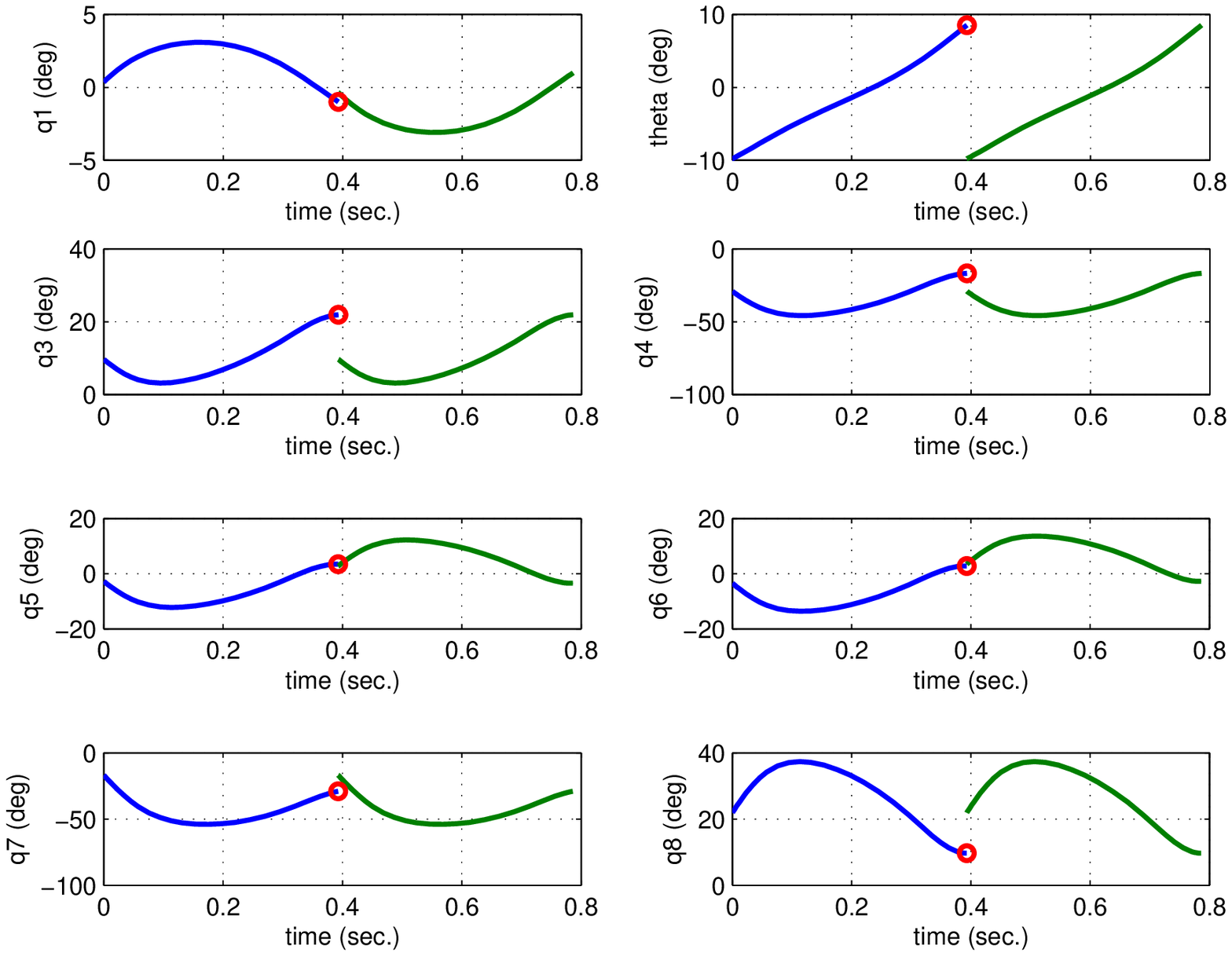}}
\caption{Joint profiles of the obtained periodic motion over
two steps, where the small circles represent $\qf^\ast $.}
\label{fig_opt_torque_pos}
\end{figure}

\begin{figure}[htbp]
\psfrag{dq1 (rad/sec)}{\footnotesize{$\dot q_1$ (rad/s)}}
\psfrag{dq2 (rad/sec)}{\footnotesize{$\dot \theta$  (rad/s)}}
\psfrag{dq3 (rad/sec)}{\footnotesize{$\dot q_3$  (rad/s)}}
\psfrag{dq4 (rad/sec)}{\footnotesize{$\dot q_4$ (rad/s)}}
\psfrag{dq5 (rad/sec)}{\footnotesize{$\dot q_5$  (rad/s)}}
\psfrag{dq6 (rad/sec)}{\footnotesize{$\dot q_6$  (rad/s)}}
\psfrag{dq7 (rad/sec)}{\footnotesize{$\dot q_7$  (rad/s)}}
\psfrag{dq8 (rad/sec)}{\footnotesize{$\dot q_8$ (rad/s)}}
\psfrag{time (sec.)}{\footnotesize{time (s)}}
\centerline{\includegraphics[width=3.in]{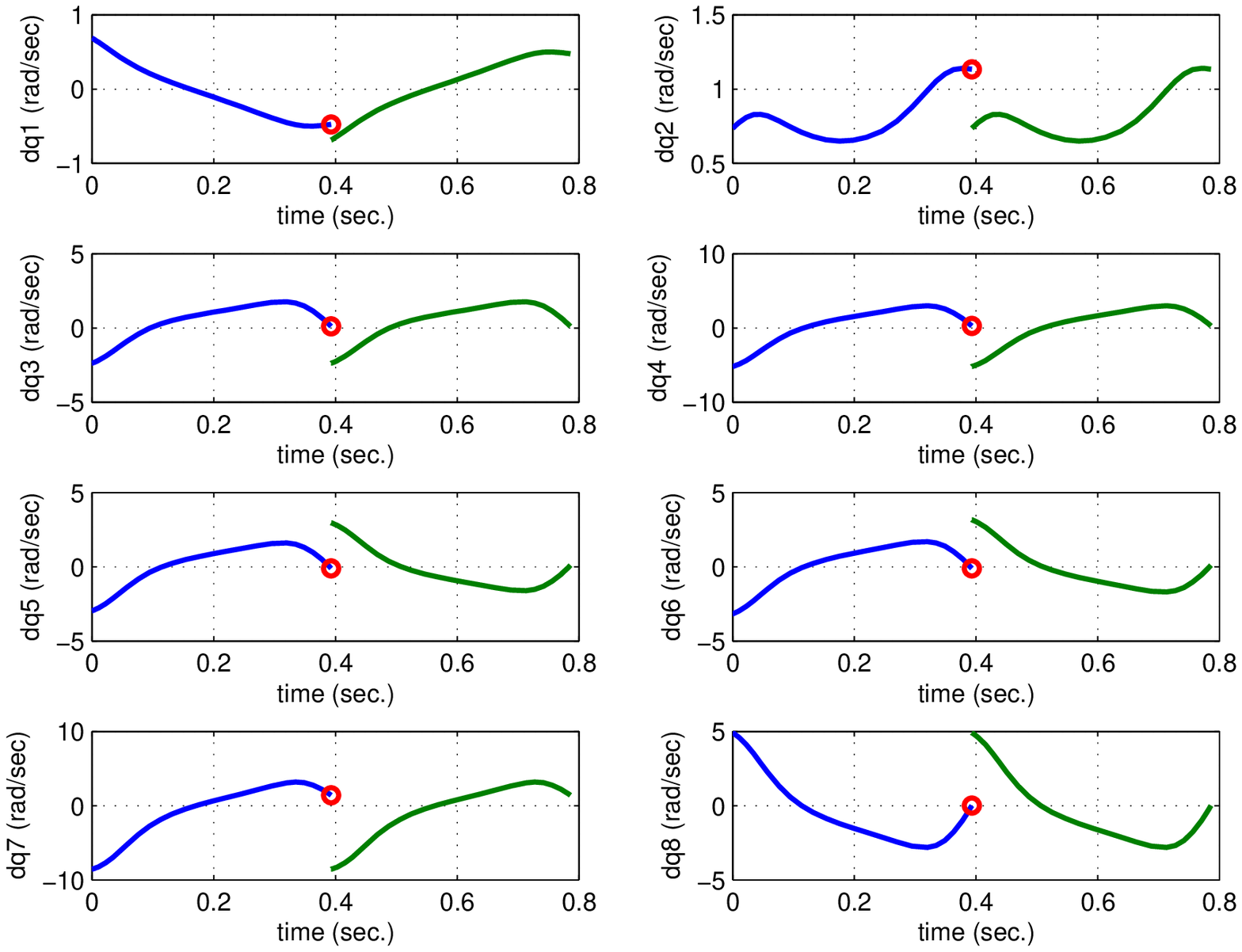}}
\caption{Joint rate profiles of the obtained periodic motion over
two steps, where the small circles represent $\dotqf^{\ast}$.}
\label{fig_opt_torque_vit}
\end{figure}

\begin{figure}[htbp]
\psfrag{tau4}{\footnotesize{u(2) (Nm)}}
\psfrag{tau3}{\footnotesize{u(1) (Nm)}}
\psfrag{tau5}{\footnotesize{u(3) (Nm)}}
\psfrag{tau6}{\footnotesize{u(4) (Nm)}}
\psfrag{tau7}{\footnotesize{u(5) (Nm)}}
\psfrag{tau8}{\footnotesize{u(6) (Nm)}}
\psfrag{time (sec.)}{\footnotesize{time (s)}}
\centerline{\includegraphics[width=3.in]{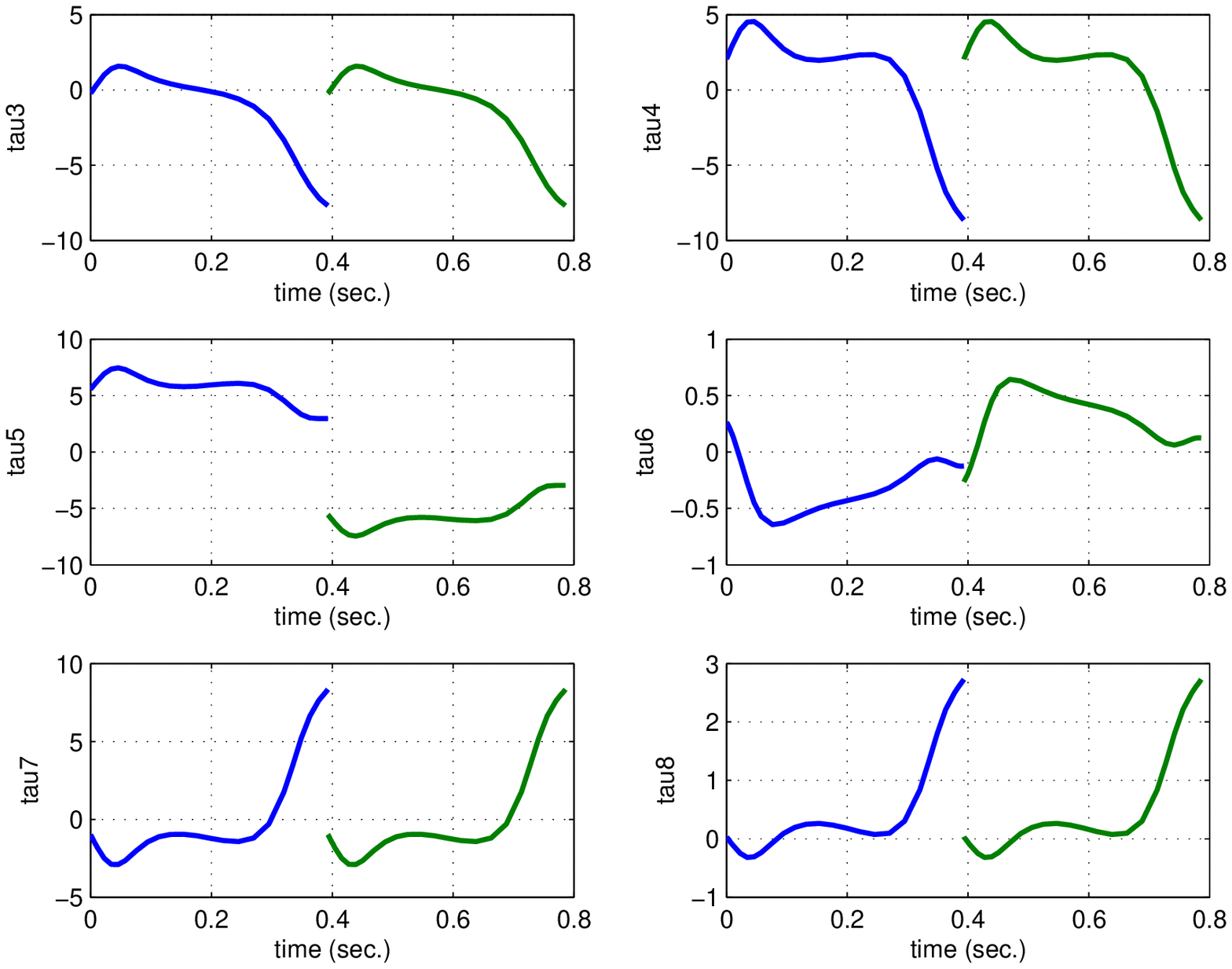}}
\caption{Torques profiles of the obtained periodic motion over
two steps}
\label{fig_opt_torque_torque}
\end{figure}

\begin{figure}[htbp]
\psfrag{F1x}{\footnotesize{$F_{st}(1)$ (N)}}
\psfrag{F1y}{\footnotesize{$F_{st}(2)$ (N)}}
\psfrag{F1z}{\footnotesize{$F_{st}(3)$ (N)}}
\psfrag{z2}{\footnotesize{$z_{sw}$ (m)}}
\psfrag{time (sec.)}{\footnotesize{time (s)}}
\centerline{\includegraphics[width=3.in]{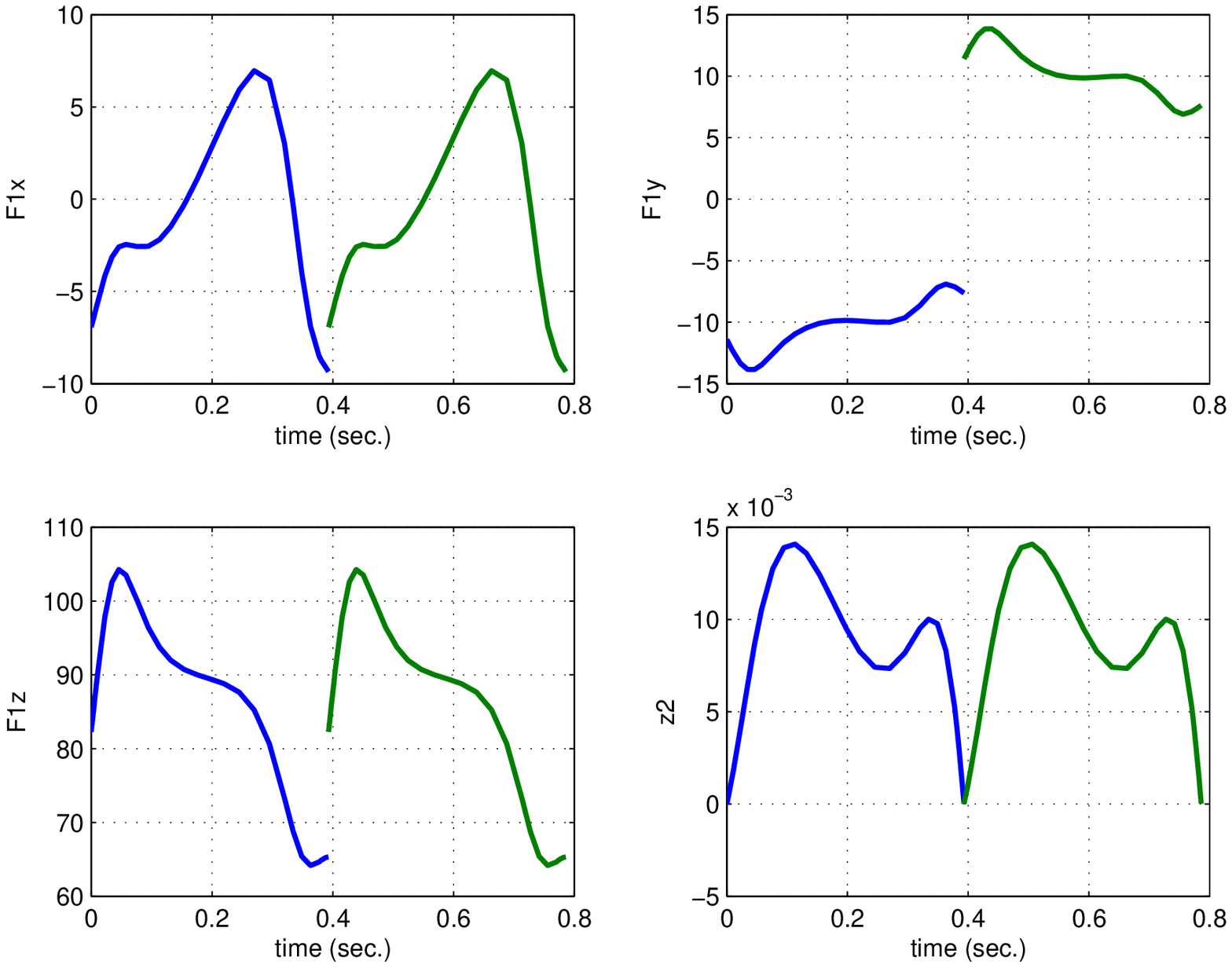}}
\caption{The reaction force on the stance foot over two
steps and the evolution of the free leg tip.}
\label{fig_opt_torque_force}
\end{figure}

A stick-figure diagram for the first step of the periodic walking gait is presented in  Fig.~\ref{fig_stick_opt_torque}. The walking gait has a period of $T=0.39$ seconds, a step size of
$L=0.176$m, and an average walking speed of $0.447$ m/sec, or $0.745$ body lengths per second. The step width is $0.156$m, close to the hip width. The
nominal gait's joint profiles and angular velocities over two consecutive steps are shown in Figs.~\ref{fig_opt_torque_pos} and \ref{fig_opt_torque_vit}, respectively. The unactuated
and actuated variables are presented; note that $\theta$ is monotonic over each step. Fig.~\ref{fig_opt_torque_torque} shows the torque required to produce the periodic motion, which is less than $10$Nm for each joint. Fig.~\ref{fig_opt_torque_force}
shows the profile of the ground reaction force on the stance foot and the profile of the swing leg tip; this figure shows that the inequality constraints are satisfied on the nominal motion.

\section{Evaluating the Stability of a Walking Cycle}
\label{sec_stability}

The stability of a fixed-point $x^\ast $ can be tested numerically
by linearizing the \Poincare\ map about the fixed-point as presented in \cite{SGC07}.
This numerical stability test has a high computational cost, however, because it requires the estimation of the Jacobian of the \Poincare\ map, in a space of dimension $2 n$-1, where $n$ is the number of independent joint coordinates; here $n=8$. We propose a slight modification of the control law in order to be able to study the stability of the closed-loop system in a reduced-dimensional state space.

\subsection{Hybrid zero dynamics (HZD) and a stability test in a reduced space}

The control law \eqref{eq_torque} is such that, on the periodic orbit, the virtual constraints \eqref{eq_constraint} are identically satisfied. However, off the periodic orbit, even if the virtual constraints are satisfied at the end of given step, they will not in general be satisfied\footnote{This may be true for several reasons, one of which is that the virtual constraints may not have been chosen to be compatible with the impact map.} at the beginning of the next step. Consequently, the behavior of the robot cannot be deduced from the behavior of the uncontrolled variables $q_u$ and the simulation of the complete model is required to predict the behavior of the robot. In the language of \cite{WEGRKO02}, \cite[Chap.~5]{WGCCM07}, while the feedback control law \eqref{eq_torque} has created a zero dynamics of the stance phase dynamics, it has not created a \textit{hybrid zero dynamics}, that is, a zero dynamics of the full hybrid model \eqref{eq_mod_hybride}.

If the control law could be modified so as to create a hybrid zero dynamics, then the study of the swing phase zero dynamics \eqref{eq_dyn_zero} and the impact model would be sufficient to determine the stability of the complete closed-loop behavior of the robot, thereby leading to a reduced-dimension stability test. A modification of the control law to achieve a hybrid zero dynamics was first proposed in \cite{MOWECHBUGR05}; a second more easily implementable method has been given in \cite{MOGR08}, along with a complete stability analysis.

Following \cite{MOGR08}, the virtual constraints are modified stride to stride so that they are compatible with the initial state of the robot at the beginning of each step. The new output for the feedback control design is
\begin{equation}
\label{eq_stride_corr}
    y_c = h(q, y_i, \dot y_i) = q_a - h_d (\theta) - h_c(\theta, y_i, \dot y_i).
\end{equation}
This output consists of the previous output \eqref{eq_constraint}, and a correction term $h_c$ that
depends on \eqref{eq_constraint} evaluated at the beginning of the step, specifically,  $y_i=
q_{a,i} - h_d (\theta_i )$ and $\dot y_i= \dot q_{a,i} - \frac{\partial h_d (\theta )}{\partial
\theta} \dot \theta_i$, where the subscript ``$i$'' denotes the initial value for the current step.
The values of $y_i, \dot y_i$ are updated at the beginning of each step and held constant
throughout the step. The function $h_c$ is taken to be a three-times continuously differentiable
function of $\theta$ such that\footnote{In our specific application, we used a fifth order
polynomial for $\theta_i \le \theta \le \frac{\theta_i+\theta_f}{2} $; continuity of position,
velocity and acceleration is ensured at $\theta = \frac{\theta_i+\theta_f}{2} $.}
\begin{equation}
\label{eq_poly_cor}
\left \{ {
    \begin{array}{rcl}
    h_c(\theta_i,y_i, \dot y_i) &=& y_i\\
    \frac{\partial h_c}{\partial \theta}(\theta_i ) &=& \frac{\dot y_i}{\dot \theta_i} \\
    h_c(\theta,y_i, \dot y_i) &\equiv& 0, \quad \frac{\theta_i+\theta_f}{2} \le \theta \le \theta_f.
    \end{array}} \right.
\end{equation}
With $h_c$ designed in this way, the initial errors of the output and its derivative are smoothly joined to the original virtual constraint at the middle of the step. In particular, for any initial error, the initial virtual constraint $h_d$ is exactly satisfied by the end of the step.

The robot's behavior with the new control law is very close to its behavior with the fixed virtual constraint, the difference being that the initial output error is zeroed by the choice of the function $h_c$ instead of being approximately zeroed by equation \eqref{eq_closed_loop}.
Our choice of $h_c$ zeroes smoothly the initial error, and the initial virtual constraints are joined at the middle of the step. Consequently, the torque required to zero the error is less than with a high-gain controller, the variation of ground reaction forces is reduced, and sliding or take-off is avoided.

Under the new control law defined by \eqref{eq_stride_corr}, the behavior of the robot is completely defined by the impact map and the swing phase zero dynamics \eqref{eq_dyn_zero}, where $h_d$ is replaced by $h_d+h_c$. The stability of a fixed-point $x^\ast $ can now be tested numerically
using a restricted \Poincare\ map defined from $S \cap Z$ to $S \cap Z$, where $Z=\{(q,\dot {q}) \vert y_c(q)=0,\; \dot y_c(q)=0 \}$ and $S$ is the switching
surface. The key point is that in $S \cap Z$, the state of the robot can be represented using only three independent variables, $x^z= [q_1, \dot q_1, \dot \theta]'$.

The restricted \Poincare\ map $P^z:S\cap Z \to S\cap Z$ induces
a discrete-time system $x^z_{k+1} =P^z(x^z_k )$. From  \cite{MOGR08}, for $\epsilon$ sufficiently small in \eqref{eq_torque}, \textit{the linearization of $P^z$ about a fixed-point determines exponential stability of the full-order closed-loop robot model.} Define $\delta x^z_k =
x^z_k -x^{z\ast} $. The \Poincare\ map linearized about a fixed-point
$x^{z\ast} =(\qoneonei^{\ast},\dotqoneonei^{\ast}, \dotthetaonei^{\ast})$ gives rise to a linearized
system,
\begin{equation}
\label{eq_lin_poin_simp}
    \delta x^z_{k+1} =A^z \delta x^z_k,
\end{equation}
where the ($3\times 3$) square matrix $A^z$ is the Jacobian of the \Poincare\
map and is computed as follows
\[
A^z=\left[ {{\begin{array}{*{20}c}
 {A^z_1 } \hfill & {A^z_2 } \hfill &  {A^z_{3} } \hfill \\
\end{array} }} \right]_{3\times 3} ,
\]
where
\begin{equation}
\label{eq_jacobian_column_i}
A^z_i =(\frac{P^z(x^{z\ast} +\Delta x^z_i )-P(x^{z\ast} -\Delta x^z_i
)}{2\Delta x^z_i }), \quad i=1,2,3,
\end{equation}
and $\Delta x^z_i =\left\{ {{\begin{array}{*{20}c}
 {\Delta q_1 }, \hfill & {i=1} \hfill \\
 {\Delta \dot {q}_1 }, \hfill & {i=2} \hfill \\
 \Delta \dot \theta,  \hfill & {i=3}. \hfill \\
\end{array} }} \right.$
The quantities $\Delta q_1$, $\Delta \dot {q}_1$, $\Delta \dot \theta$ are small perturbations introduced to calculate the linearized model, and the denominator in \eqref{eq_jacobian_column_i} must of course be interpreted as the scalar perturbation used in computing $\Delta x^z_i$. In general, the calculation of the Jacobian is sensitive to the amplitude of the perturbation $\Delta x^z_i$.
The calculation of matrix $A^z$ requires six evaluations of the function $P^z(x^{z\ast} \pm \Delta x^z_i )$. Each evaluation of this function is composed of the reconstruction of the vector $x_{1}$ from $x^z$ with $x^z \in S\cap Z$, the calculation of the impact map \eqref{eq_imp_pos}-\eqref{eq_imp_vit}, the calculation of $y_i, \dot y_i$, the calculation of $h_c(\theta)$ and the integration of the swing phase zero dynamics. A fixed-point of the restricted \Poincare\ map is locally exponentially stable, if, and only if, the eigenvalues of $A^z$ have magnitude strictly less than one \cite[Chap.~4]{WGCCM07}.

\subsection{Example of the periodic motion minimizing integral-squared torque}
\label{sec:FirstExampleController}

We consider the virtual constraints corresponding to the optimal periodic motion obtained in Section \ref{sec_opt_motion_torque}, and the
control law defined by \eqref{eq_stride_corr} is used.

To study the stability of this control law around the periodic motion,
the eigenvalues of the linearized restricted \Poincare\ map are computed (for $\Delta q_i=0.0750 ^\circ$, $\Delta \dot q_i=\Delta \dot \theta=0.375 ^\circ s^{-1}$), yielding
\begin{equation}
\label{eq_Az}
    A^z=\left[{ \begin{array}{ccc}
    0.1979  & -0.4625 &  -0.2145\\
    5.8899 &  -2.8417 &  -1.7476\\
   -4.7411  & -0.2132  &  0.7809
    \end{array}} \right ].
\end{equation}
The three eigenvalues are:
\begin{eqnarray*}
    \lambda_1 &=&  0.8878  \\
  \lambda_2 &=& -0.6951  \\
  \lambda_3 &=& -2.0891
\end{eqnarray*}
One eigenvalue has magnitude greater than one and hence the gait is unstable under this controller.

We have found that for most periodic motions optimized with respect to integral-squared torque per step length, \eqref{opt_criterion}, the obtained gait is unstable under the control law defined by \eqref{eq_stride_corr}. In the following sections, three strategies to obtain stable walking will be considered. In the first strategy, closed-loop stability is directly considered in the design of the periodic motion. In the second strategy, a stride-to-stride controller is introduced to stabilize the gait. In the third strategy, freedom in the selection of the controlled outputs\footnote{The controlled outputs are no longer the actuated variables as in \eqref{eq_torque}, but a judiciously chosen linear combination of $q$. A convenient choice of outputs is given.} is used to obtain a stable walking cycle using only within-stride control.

\section{Periodic motion optimized with respect to a stability criterion}

Because we are using optimization to compute a periodic solution of the model, it is possible to consider the stability condition (i.e., magnitude of eigenvalues less than one) either as a criterion in the optimization process, or as a constraint.

\begin{figure}[htbp]
\centerline{\includegraphics[width=3.in]{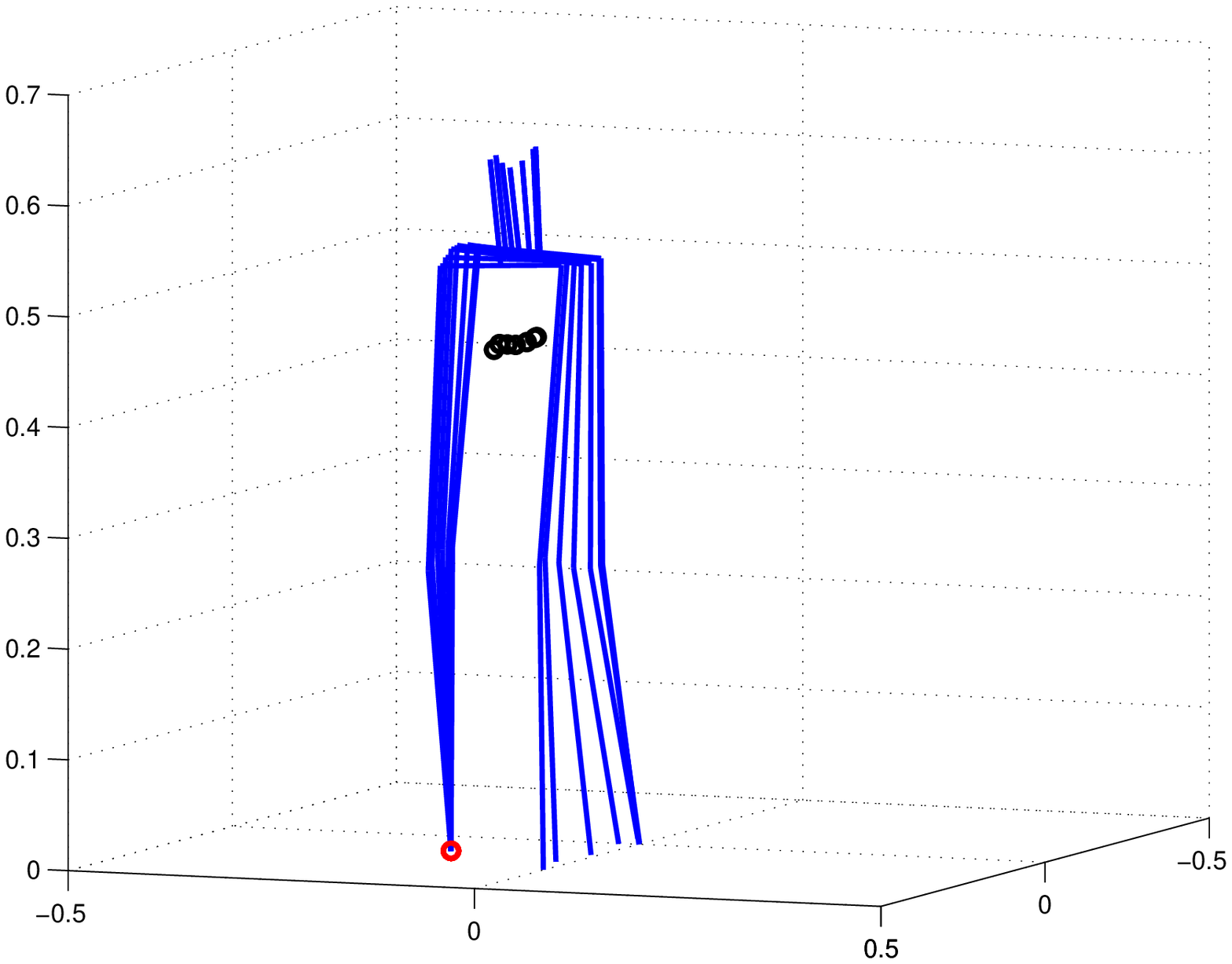}}
\caption{Stick-diagram of the optimal motion with respect to the stability criterion}
\label{fig_stick_opt_stab}
\end{figure}

\begin{figure}[htbp]
\psfrag{q1 (deg)}{\footnotesize{$q_1$ (deg)}}
\psfrag{theta (deg)}{\footnotesize{$\theta$  (deg)}}
\psfrag{q3 (deg)}{\footnotesize{$q_3$ (deg)}}
\psfrag{q4 (deg)}{\footnotesize{$q_4$ (deg)}}
\psfrag{q5 (deg)}{\footnotesize{$q_5$ (deg)}}
\psfrag{q6 (deg)}{\footnotesize{$q_6$ (deg)}}
\psfrag{q7 (deg)}{\footnotesize{$q_7$ (deg)}}
\psfrag{q8 (deg)}{\footnotesize{$q_8$ (deg)}}
\psfrag{time (sec.)}{\footnotesize{time (s)}}
\centerline{\includegraphics[width=3.in]{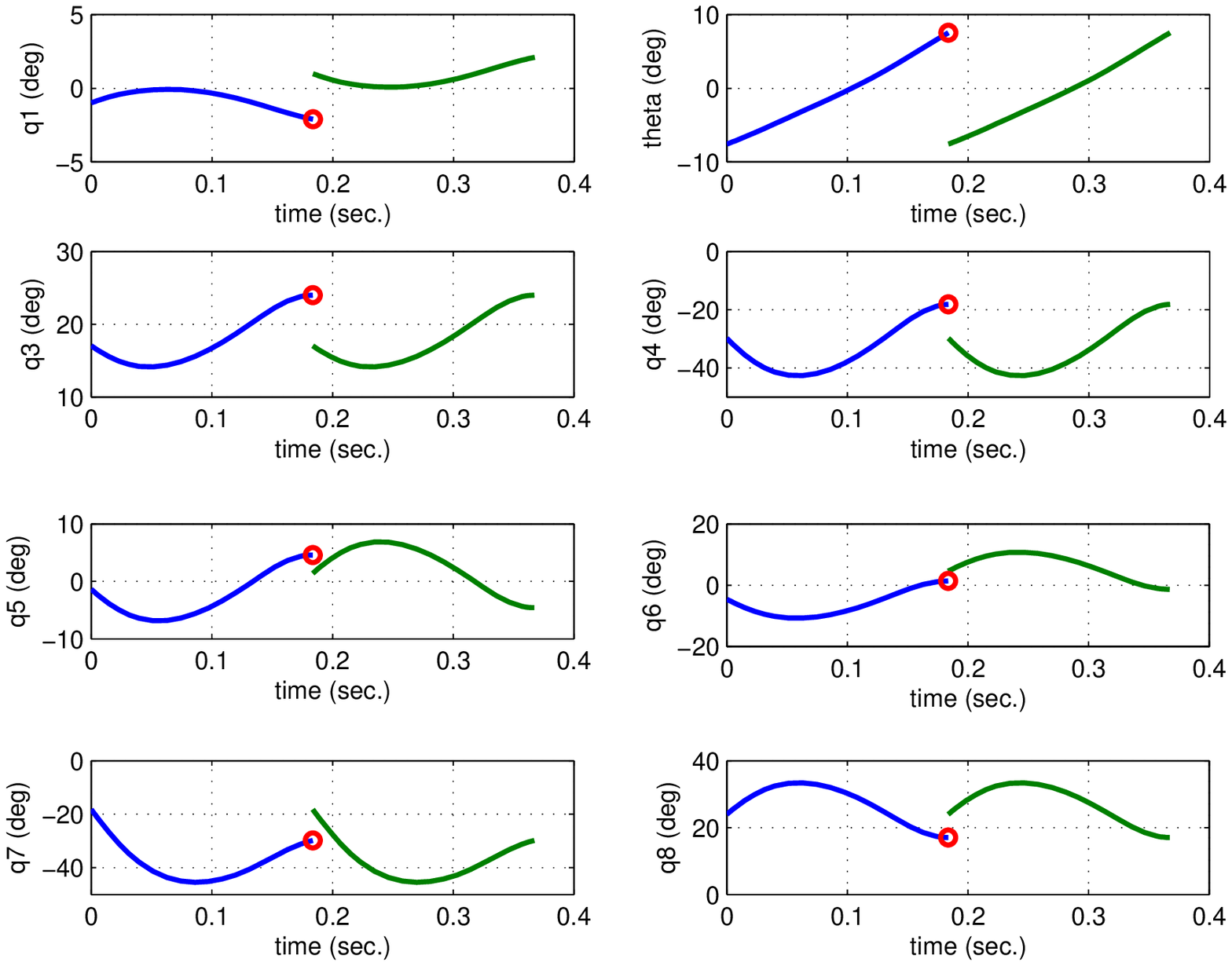}}
\caption{Joint profiles of the obtained periodic motion over
two steps, where the small circles represent $\qf^\ast $.}
\label{fig_opt_stab_pos}
\end{figure}

\begin{figure}[htbp]
\psfrag{dq1 (rad/sec)}{\footnotesize{$\dot q_1$ (rad/s)}}
\psfrag{dq2 (rad/sec)}{\footnotesize{$\dot \theta$  (rad/s)}}
\psfrag{dq3 (rad/sec)}{\footnotesize{$\dot q_3$  (rad/s)}}
\psfrag{dq4 (rad/sec)}{\footnotesize{$\dot q_4$ (rad/s)}}
\psfrag{dq5 (rad/sec)}{\footnotesize{$\dot q_5$  (rad/s)}}
\psfrag{dq6 (rad/sec)}{\footnotesize{$\dot q_6$  (rad/s)}}
\psfrag{dq7 (rad/sec)}{\footnotesize{$\dot q_7$  (rad/s)}}
\psfrag{dq8 (rad/sec)}{\footnotesize{$\dot q_8$ (rad/s)}}
\psfrag{time (sec.)}{\footnotesize{time (s)}}
\centerline{\includegraphics[width=3.in]{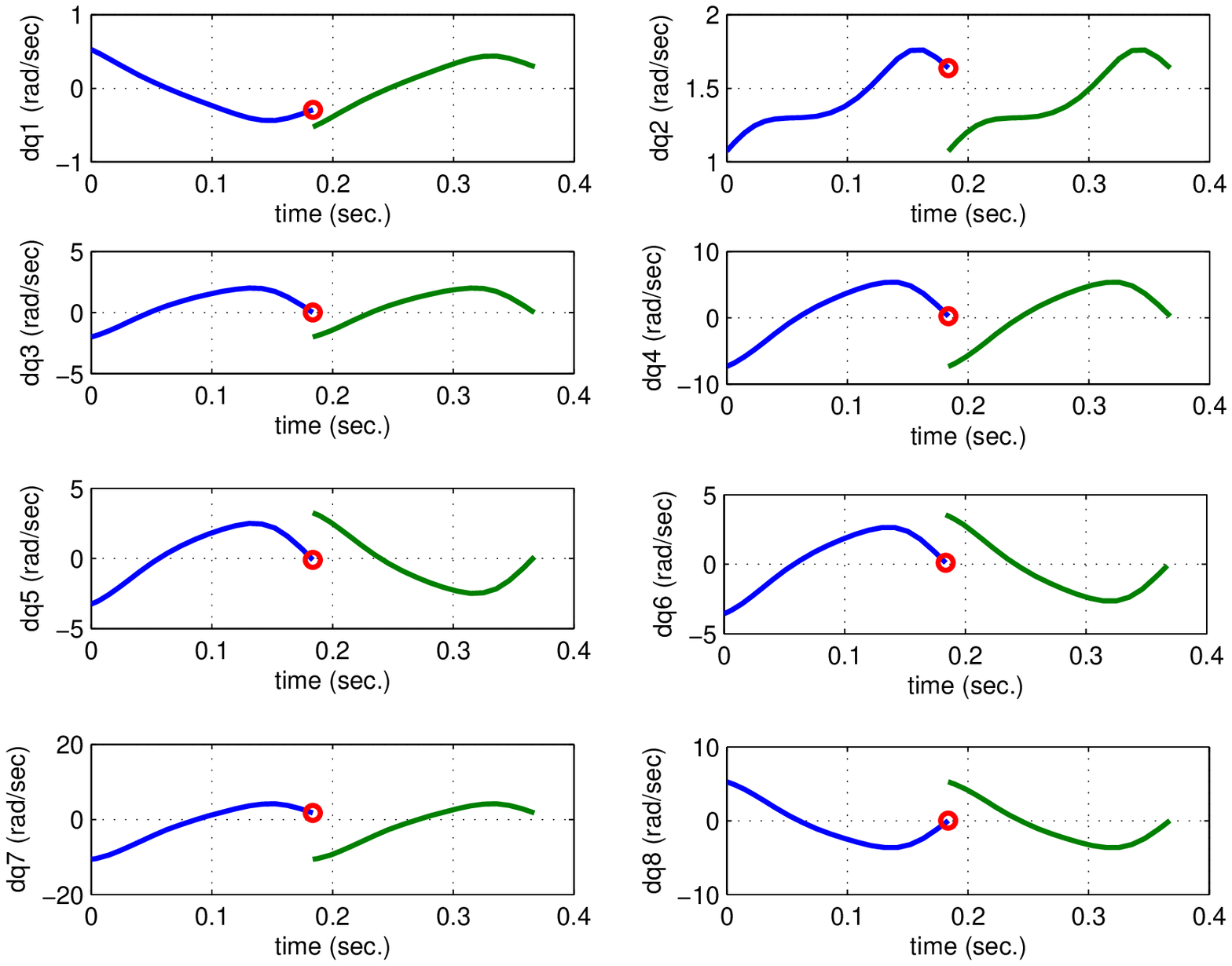}}
\caption{Joint rate profiles of the obtained periodic motion over
two steps, where the small circles represent $\dotqf^{\ast}$.}
\label{fig_opt_stab_vit}
\end{figure}

Using the stability test above, which has a reasonable calculation cost, the selection of a periodic motion that minimizes the maximum of the magnitudes of the eigenvalue of $A^z$ was performed,
$$J=\max \{ |\lambda(A^z)| \}, $$
where $\lambda(A^z)$ is the set of eigenvalues of $A^z$.
The optimization process is similar to the optimization process described in Section \ref{sec:optimization}, only the criterion is changed. Starting from the same initialization for the optimization process, a new periodic trajectory is obtained. This trajectory is defined by
\[\begin{array}{c}\qf^\ast = [
-0.0306,   -0.3304,    0.3892,   -0.2853,    0.0703,    0.0265\\   -0.4948,    0.2827]'
\\
\dotqf^\ast =[
   -0.2719,   -1.6158,   -0.0710,   -0.1553,   -0.1998,    0.2312 \\   1.1816,  -0.0450 ]'.
\end{array}
\]
The stick-diagram for a step is presented in Fig.~\ref{fig_stick_opt_stab}. The walking cycle has a period of $T=0.175$ seconds, a step length of
$L=0.144$ m, and an average walking speed of $0.82$ m/sec. The step width is $0.173$ m, close to the hip width. The
nominal gait's joint profiles and angular velocities over two
steps are shown in Figs.~\ref{fig_opt_stab_pos} and \ref{fig_opt_stab_vit}, respectively. The unactuated and actuated variables are presented, and it is shown that $\theta$ is monotonic.
The duration of a step is shorter than in the case of torque optimization (i.e., the motion is faster). The movement in the frontal plane has smaller amplitude; indeed, the maximal side-to-side variation of the center of mass in the frontal direction is less than $1.5$ cm (see Fig.~\ref{fig_stick_opt_stab}), while this deviation was more than $5$ cm in the case of torque optimization (see Fig.~\ref{fig_stick_opt_torque}).

The eigenvalues of the matrix $A^z$ are
\begin{eqnarray*}
    \lambda_{1,2}  &=&  -0.873 \pm  216 i  \\
  |\lambda_{1,2}|  &=&  0.890    \\
  \lambda_3  &=&  0.887
\end{eqnarray*}
All of the eigenvalues have magnitude less than 1.0, indicating that
the obtained nominal orbit $x^\ast $ is locally exponentially
stable. To illustrate the orbit's local stability, the 3D biped's full model in
closed-loop is simulated with an initial state perturbed from the fixed-point
$x^\ast $. An initial error of $-0.5^\circ$ is introduced on each joint.

\begin{figure}[htbp]
\psfrag{q1minus (rad)}{\small{$\qoneonef$ (rad)}}
\psfrag{thminus (rad)}{\small{$\thetaonef$  (rad)}}
\psfrag{dq1minus (rad/sec)}{\small{$\dotqoneonef$  (rad/s)}}
\psfrag{dthminus (rad/sec)}{\small{$\dotthetaonef$ (rad/s)}}
\psfrag{step number}{\small{step number}}
\centerline{\includegraphics[width=3.in]{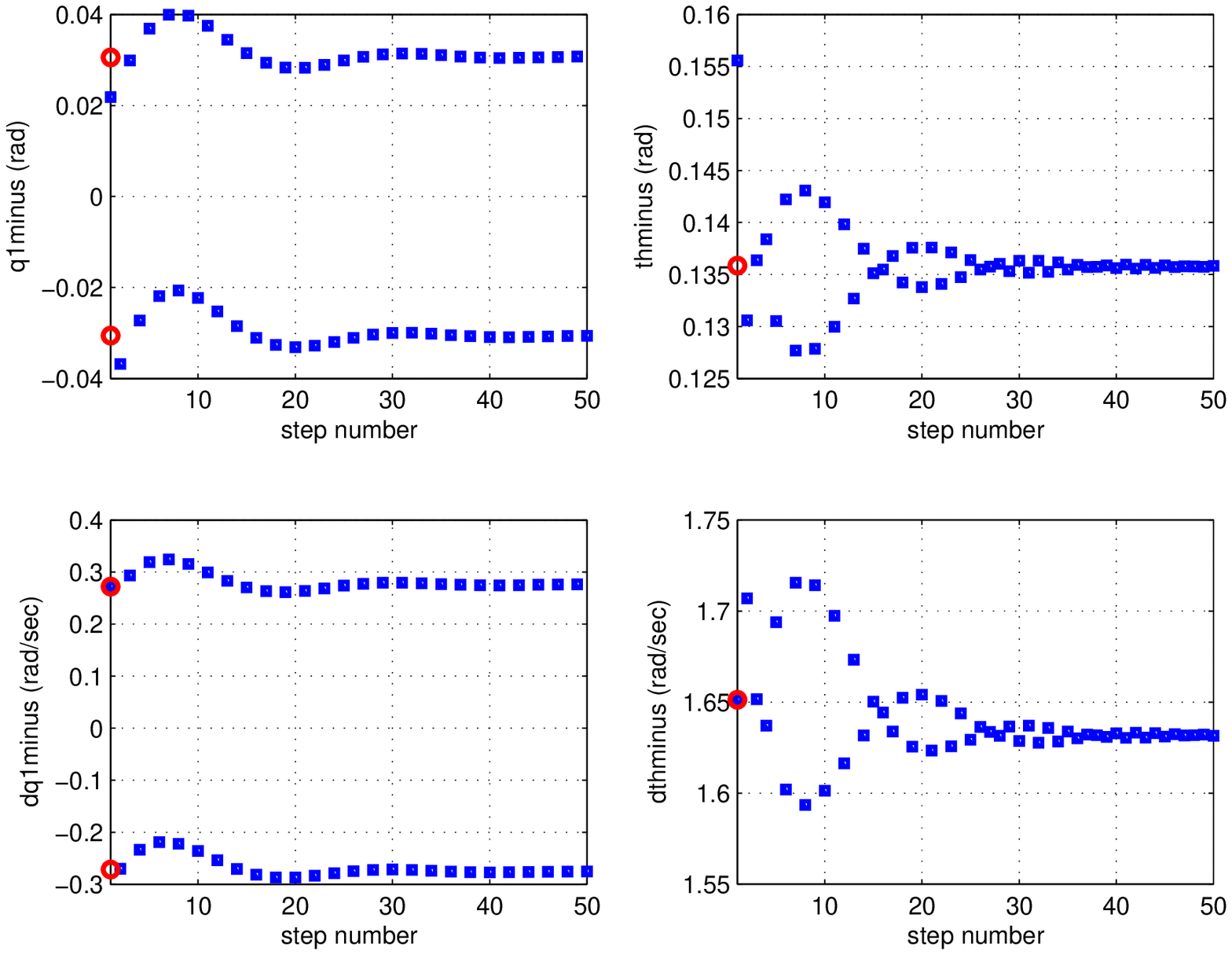}}
\caption{The evolution of $q_u$ at the end of each step for the
3D biped's full model under closed-loop walking control,
with the initial condition perturbed from $x^\ast$. The small circles represent the values on the periodic orbit.}
\label{fig_control_eigen_qu}
\end{figure}

Fig.~\ref{fig_control_eigen_qu} shows the evolution of the final values of the uncontrolled
variables $q_u$ from one step to the next. These variables converge toward the periodic motion.
With the modification of the virtual constraints by the introduction of the polynomial $h_c$, the
output $y_c$ and its derivative are zero throughout each step.

\begin{figure}[htbp]
\psfrag{q1 (rad)}{\small{$\dot q_1$  (rad/s)}}
\psfrag{qp1 (rad/s)}{\small{$q_1$ (rad)}}
\psfrag{q2 (rad)}{\small{$\dot q_2$  (rad/s)}}
\psfrag{qp2 (rad/s)}{\small{$q_2$ (rad)}}
\psfrag{q3 (rad)}{\small{$\dot q_3$  (rad/s)}}
\psfrag{qp3 (rad/s)}{\small{$q_3$ (rad)}}
\psfrag{q4 (rad)}{\small{$\dot q_4$  (rad/s)}}
\psfrag{qp4 (rad/s)}{\small{$q_4$ (rad)}}
\centerline{\includegraphics[width=3.in]{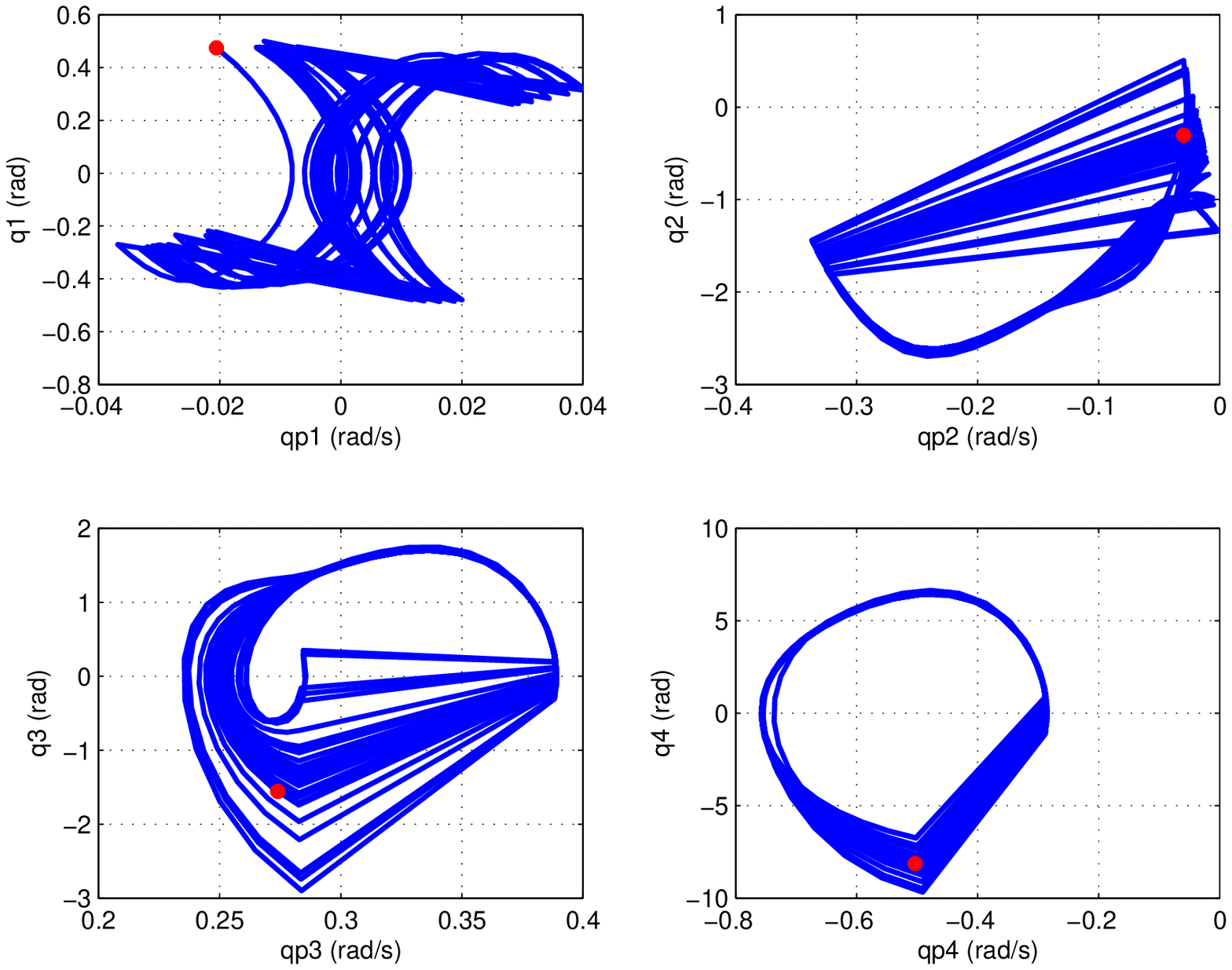}}
\caption{Phase-plane plots for $q_i$, $i=1,\ldots, 4$. The straight lines correspond to the impact phase, where the state of the robot changes instantaneously. The initial state is represented by a (red) star. Each variable converges to a periodic motion.}
\label{fig_control_eigen_state}
\end{figure}

Fig.~\ref{fig_control_eigen_state} shows the phase-plane evolution of the first four variables as an illustration of the behavior of the state of the robot. The discontinuity at impact manifests itself as a straight line on the plots. The desired motion of the robot is periodic, with exchange of the legs at impact. This behavior corresponds to a motion with a period of one step for the variables $q_2$, $q_3$ and $q_4$ and a period of two steps for $q_1$; see \eqref{eq_symetrie}. The convergence towards a periodic motion is clear for both the controlled and uncontrolled variables.

\section{Stride-to-Stride controller}

If a desired periodic gait is not stable,
or if the corresponding rate of convergence is not sufficiently
rapid, then event-based control may be designed and integrated with the
continuous, stance phase controller \cite{GRIZZLEJ06}.

\subsection{General method}
Let $\beta$ be a
vector of parameters that are held constant during the stance phase
and updated at each impact. The parameters could be a subset of the
parameters used to specify the virtual constraints, $\alpha _0
,\cdots ,\alpha _3 $, or an auxiliary set of parameters. Moreover, the parameter updates can be done on the basis of the full-state of the robot as in \cite{SGC07}, or on the basis of the state of the hybrid zero dynamics, which is done here.

The output in \eqref{eq_stride_corr} is augmented with an additional term $h_s (\theta,\beta)$ depending on a vector of parameters $\beta$:
\begin{equation}
\label{eq_stride_beta}
    y = h(q, y_i, \dot y_i,\beta) = q_a - h_d (\theta) - h_c(\theta, y_i, \dot y_i)- h_s (\theta,\beta)
\end{equation}
with\footnote{In our specific application, we used a sixth order polynomial for $\theta_i \le \theta \le 0.1 \theta_i + 0.9 \theta_f $. Continuity of position, velocity and acceleration is ensured at $\theta = 0.1 \theta_i + 0.9 \theta_f $.}
\begin{equation}
\left \{ {
    \begin{array}{rcl}
    h_s(\theta_i,\beta) &=& 0\\
    \frac{\partial h_s}{\partial \theta}(\theta_i, \beta ) &=& 0 \\
    h_s(\frac{\theta_i+ \theta_f}{2},\beta) &=& \beta\\
    h_s(\theta,\beta) &\equiv& 0, \quad 0.1 \theta_i+0.9 \theta_f \le \theta \le \theta_f.
    \end{array}} \right.
\end{equation}
This choice is convenient because it allows the value of the virtual constraints at mid-stance to be updated, without requiring a re-design of the event-based controller that created the hybrid zero dynamics. As long as the impact occurs for $\theta \ge 0.1 \theta_i + 0.9 \theta_f$, which happens when the motion is close to the periodic orbit, the final state of the robot is on the original hybrid zero dynamics independently of any modification $h_s$ for the previous step.

The \Poincare\ map can now be viewed as a
nonlinear control system on $S \cap Z$ with inputs $\beta_k $, where $\beta_k $ is the value of $\beta$ during the step $k+1$, namely $x^z_{k+1}
=P(x^z_k ,\beta _k )$. Linearizing this nonlinear system about the
fixed point $x^{z\ast}$ and the nominal
parameter value $\beta ^\ast = 0_{6 \times 1} $ leads to
\begin{equation}
\label{eq_ponc_stride}
    \delta x^z_{k+1} = A^z \delta x^z_k + F \delta \beta _k ,
\end{equation}
where $\delta \beta_k =\beta_k -\beta ^\ast $ and $F$ is the
Jacobian of $P$ with respect to $\beta $. Designing a feedback matrix
\begin{equation}
\label{eq_K_DLQR}
\delta \beta_k =-K \delta  x^z_k
\end{equation}
such that the
eigenvalues of $(A^z-FK)$ have magnitude strictly less than one will
exponentially stabilize the fixed point $x^{\ast} $.

\subsection{Example of the periodic motion minimizing the integral-squared torque}

The periodic motion described in Section \ref{sec_opt_motion_torque} can be stabilized using a stride-to-stride controller.
The virtual constraints are modified according to equation (\ref{eq_stride_beta}); for the $k$-th step the virtual constraints $h_{d,k}$ are
\begin{equation}
\label{eq_stride_virt_const}
    h_{d,k} = h_d (\theta) - h_c(\theta, y_i, \dot y_i)- h_s (\theta,\beta_k).
\end{equation}
The $(3 \times 6)$ matrix $F$ was computed numerically, analogously to \eqref{eq_lin_poin_simp}, yielding
\begin{equation}
\label{eq_F}
    \left[{ \begin{array}{cccccc}
  -0.030 &  -0.028  &  0.141 &  -0.065   & 0.028  &  0.018\\
   -0.237 &  -0.233 &   1.073 &  -0.494  &  0.225 &   0.144\\
    0.163 &   0.224 &  -0.023 &   0.001  & -0.231 &  -0.064\\
                \end{array}} \right ]
\end{equation}
The matrix $A^z$ is given in \eqref{eq_Az}.

The $(6 \times 3)$ gain matrix $K$ was calculated via DLQR so that the state-feedback law  $\delta \beta_k =-K \delta  x^z_k$  minimized the cost function $\sum_k {( \delta  x^{z'}_k\delta  x^z_k + r \delta \beta_k' \delta \beta_k)}$, subject to the state dynamics \eqref{eq_ponc_stride}. The coefficient $r$ allows a tradeoff to be made between the convergence rate and the amplitude of the change of the virtual constraint, which affects the region of convergence. For $r=2$ we obtain
\begin{equation}
\label{eq_K}
    K=\left[{ \begin{array}{ccc}
-0.603  &  0.243 &   0.176\\
   -0.607  &  0.186 &   0.171\\
    2.566  & -1.704 &  -0.814\\
   -1.181  &  0.793 &   0.376\\
    0.590  & -0.164 &  -0.165\\
    0.359  & -0.179 &  -0.108\\ \end{array}} \right ].
\end{equation}

The eigenvalues of the linearized \Poincare\ map in closed loop $A^z - F K$ indicate that the gait is stable with the addition of the stride-to-stride controller\footnote{The controller of Sec.~\ref{sec:FirstExampleController} is used within stride and the event-based controller is applied at each impact.}. The eigenvalues become
\begin{eqnarray*}
    \lambda_1 &=&   0.7906  \\
  \lambda_{2,3} &=& -0.4478 \pm 0.047 i    \\
  |\lambda_{2,3}| &=&  0.4503
\end{eqnarray*}

To illustrate the orbit's local stability, the 3D biped's complete model in
closed-loop is simulated with the initial state perturbed away from the fixed-point
$x^\ast $. An initial error of $-1^\circ$ is introduced on each joint and a velocity error of $-5^\circ s^{-1}$ is introduced on each joint velocity.

\begin{figure}[htbp]
\psfrag{q1minus (rad)}{\small{$\qoneonef$ (rad)}}
\psfrag{thminus (rad)}{\small{$\thetaonef$  (rad)}}
\psfrag{dq1minus (rad/sec)}{\small{$\dotqoneonef$  (rad/s)}}
\psfrag{dthminus (rad/sec)}{\small{$\dotthetaonef$ (rad/s)}}
\psfrag{step number}{\small{step number}}
\centerline{\includegraphics[width=3.in]{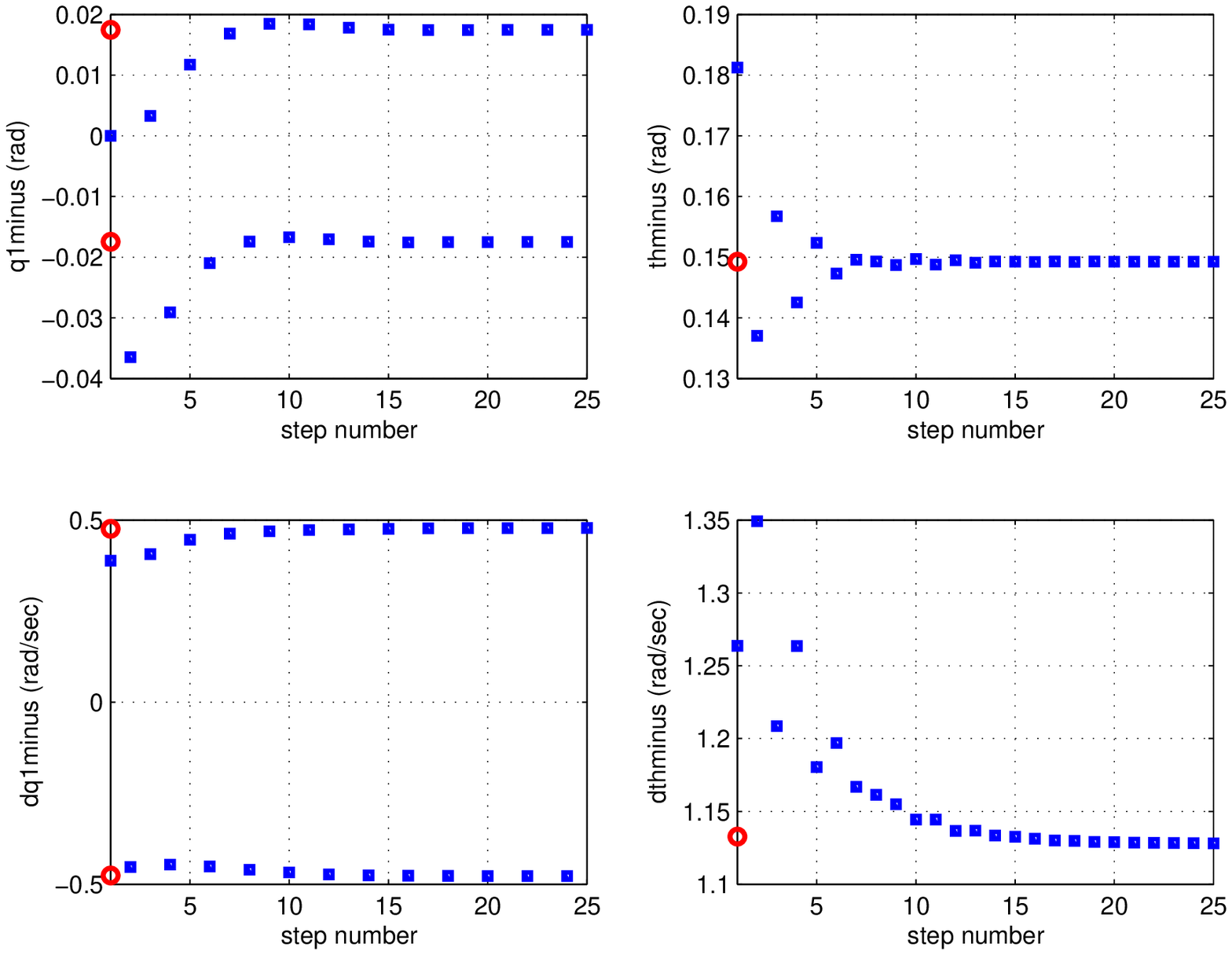}}
\caption{The evolution of $q_u$ at the end of each step for the
3D biped's full model under closed-loop walking control,
with the initial condition perturbed from $x^\ast$. The small circles represent the values on the periodic orbit.}
\label{fig_control_stride_qu}
\end{figure}

Fig.~\ref{fig_control_stride_qu} shows the
evolution of the values of the uncontrolled variables $q_u$ at the end of each step, when event-based DLQR controller is used. These variables clearly converge toward the periodic motion.

\begin{figure}[htbp]
\psfrag{q1 (rad)}{\small{$\dot q_1$  (rad/s)}}
\psfrag{qp1 (rad/s)}{\small{$q_1$ (rad)}}
\psfrag{q2 (rad)}{\small{$\dot q_2$  (rad/s)}}
\psfrag{qp2 (rad/s)}{\small{$q_2$ (rad)}}
\psfrag{q3 (rad)}{\small{$\dot q_3$  (rad/s)}}
\psfrag{qp3 (rad/s)}{\small{$q_3$ (rad)}}
\psfrag{q4 (rad)}{\small{$\dot q_4$  (rad/s)}}
\psfrag{qp4 (rad/s)}{\small{$q_4$ (rad)}}\centerline{\includegraphics[width=3.in]{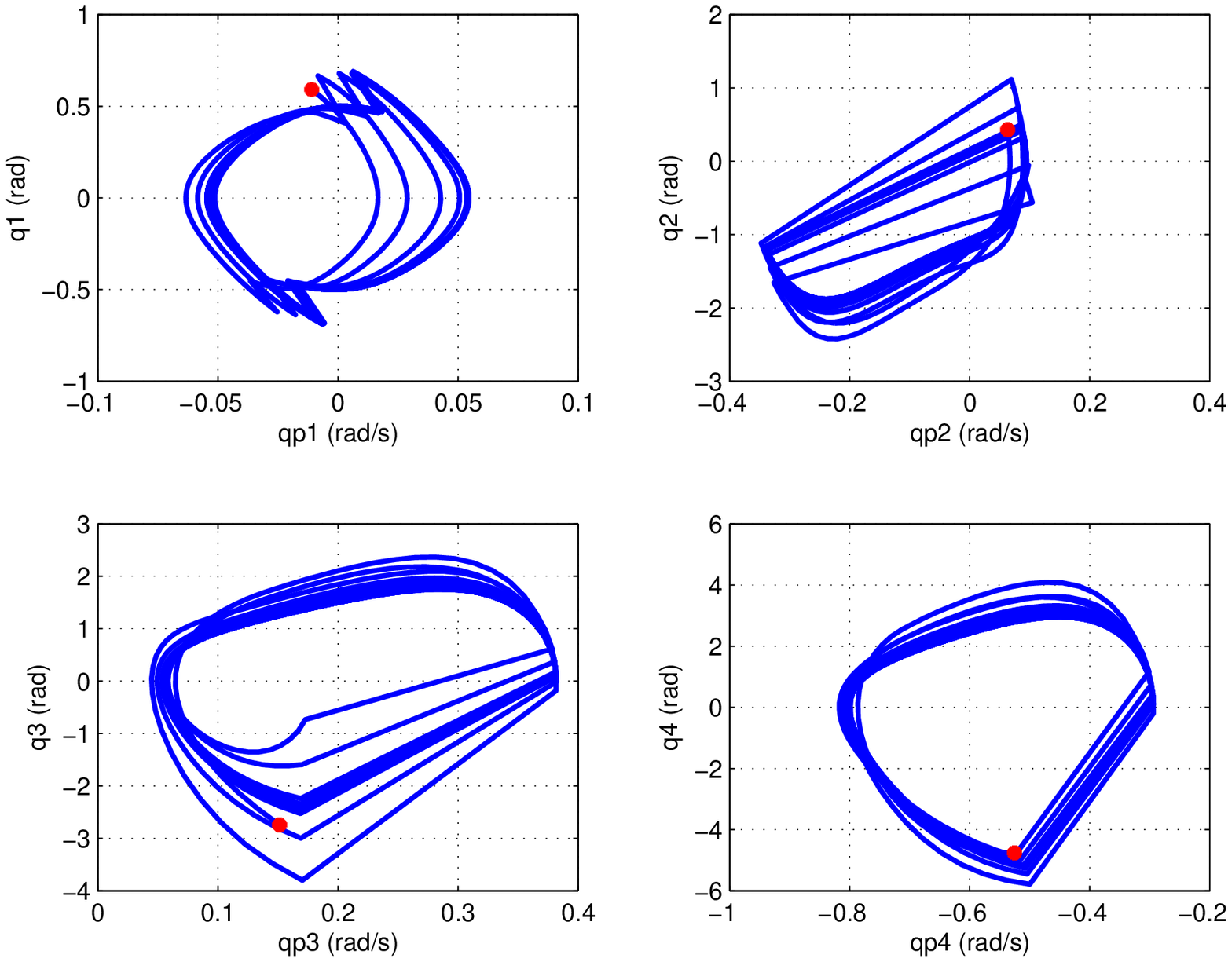}}
\caption{Phase-plane plots for $q_i$, $i=1,\ldots, 4$. The straight lines correspond to the impact phase, where the state of the robot changes instantaneously. The initial state is represented by a (red) star. Each variable converges to a periodic motion.}
\label{fig_control_stride_state}
\end{figure}

Fig.~\ref{fig_control_stride_state} shows phase-plane plots of the first four variables. The convergence towards a periodic motion is clear for the controlled and uncontrolled variables.

\section{Improved Selection of the Outputs to be Controlled}
\label{sec:improved_outputs}

In the previous two sections, the controlled variables driven by the virtual constraints are simply
the actuated variables, $q_a$; see \eqref{eq_constraint}. The choice of the controlled variables
directly affects the zero dynamics in \eqref{eq_dyn_zero}. It is shown here that for the same
desired periodic motion, the stability of the closed-loop system can be dramatically improved
through a judicious choice of the controlled variables.

\subsection{Effect on the swing phase zero dynamics}
For simplicity, we limit our analysis to the case of controlled variables that are linear with respect to the configuration variables. Thus the controlled variables are
\begin{equation}
\label{eq:MaM1}
    q_c = M \left[ {\begin{array}{c}
 {q_1 } \\
 \theta \\
 {q_a } \\
\end{array} } \right] = \left[{\begin{array}{ccc} M_1 &M_{\theta}& M_a\end{array} } \right] \left[ {\begin{array}{c}
 {q_1 } \\
 \theta \\
 {q_a } \\
\end{array} } \right],
\end{equation}
where $M$ is a $(6 \times 8)$ constant matrix with $M_a$ invertible.
A known periodic motion $q^*(t)$ can be reparameterized\footnote{This assumes that $\theta$ is monotonic.} as function of the variable $\theta$, yielding $q^*(\theta)$. The virtual constraint for the new controlled variables then yields the output
\begin{equation}
\label{eq_output}
    y = \underbrace{ M \left[ {\begin{array}{c}
 {q_1 } \\
 \theta \\
 {q_a } \\
\end{array} } \right]}_{h_0(q)} - \underbrace{M \left[ {\begin{array}{c}
 q_1^*(\theta) \\
 \theta \\
 q_a^*(\theta) \\
\end{array} } \right]. }_{h_d(\theta)}
\end{equation}

When the constraint is satisfied, $y \equiv 0$, equation \eqref{eq_output} allows us to solve for $q_a$, giving
\begin{equation}
\label{eq_qa_output}
    q_a = q_a^*(\theta) + M_a^{-1}M_1 \left( q_1^*(\theta)-q_1 \right).
\end{equation}
Substituting this equation into \eqref{eq_mod_dyn_un},
we obtain for the swing phase zero dynamics
\begin{equation}
\label{eq_dyn_zero_output}
\begin{array}{c} D_{11}(q_u)\left[ {{\begin{array}{c}
 {\ddot q_1 } \\
 {\ddot \theta } \\
\end{array} }} \right] + D_{12}(q_u)\left({ \frac{\partial \,q_a^* }{\partial
\,\theta } \ddot \theta + {\frac{\partial ^2 q_a^*}{\partial \,\theta^2 }} \dot \theta^2} \right) +\\D_{12}(q_u)M_a^{-1}M_1 \left({ \frac{\partial \,q_1^* }{\partial
\,\theta } \ddot \theta + {\frac{\partial ^2 q_1^* }{\partial \,\theta^2 }} \dot \theta^2-\ddot q_1} \right)+H_1(q_u, \dot q_u) =0.
\end{array}
\end{equation}

The nominal periodic motion satisfies both equations \eqref{eq_dyn_zero_output} and \eqref{eq_dyn_zero}, but the two equations produce different solutions away from the periodic motion.
When the principle of virtual constraints is applied to a system with only one degree of underactuation, namely $\theta$, which is common for example in planar bipeds, the swing phase zero dynamic is not affected by the choice of the output, and therefore the stability of a periodic orbit (i.e., walking motion) is not modified; only the transient motion can be different. In other words, for planar robots with one degree of underactuation, the stability depends only on the trajectory of the periodic orbit and not on the choice of virtual constraints used to achieve it \cite{CHFODJ03}, \cite[pp.~160]{WGCCM07}.

In the case of a system with two degrees of underactuation, the choice of the controlled output can affect the stability of the gait via the choice of $M_a^{-1}M_1$.
In order to illustrate this property, a new choice of output is proposed. This choice is based on the following physical reasoning: The motion in the frontal direction is difficult to stabilize. The position of the center of mass in the frontal direction is important. If, at touchdown, the center of mass is not between the feet, but outside the position of the next supporting foot, the robot will topple sideways. Thus, the control of the variable $q_6$ (which regulates step width on the swing leg) is replaced by the control of the distance between the swing leg end and the center of mass along the frontal direction. To obtain a linear output, this function is linearized around the touchdown configuration to define $M$ in \eqref{eq_output}.

\subsection{Example of the periodic motion minimizing integral-squared torque}

The periodic motion described in Section \ref{sec_opt_motion_torque} can be stabilized using the new controlled output.
As mentioned in the previous subsection, the actuated joints $q_3$, $q_4$, $q_5$, $q_7$, and $q_8$ are controlled via virtual constraints just as in the original control law. A new output $h_{d,4}$ is considered, with this output no longer based on $q_6$ but instead on distance between the swing leg end and the center of mass along the frontal direction.

For this trajectory, for support on leg 1, the linearization around $\qf$ of the distance between the swing leg end and the center of mass along the frontal direction yields
    \[\begin{array}{c}d=-0.457 q_1-0.020 q_2-0.018 q_3-0.020 q_4 - 0.489 q_5 \\+ 0.461 q_6-0.056 q_7 - 0.022 q_8.\end{array}
\]
On the periodic orbit, this distance is evaluated and approximated by a function of $\theta$, denoted $d^*(\theta)$. The new controlled output is then
\begin{equation}
    \begin{array}{c}y_4=-0.457 q_1-0.020 q_2-0.018 q_3-0.020 q_4 - 0.489 q_5 \\+ 0.461 q_6-0.056 q_7 - 0.022 q_8 - d^*(\theta).\end{array}
\end{equation}

When the control law is defined using this new output, the walking gait is stable, as can be shown via the calculation of the eigenvalues of the linearization of the restricted \Poincare\ map, $A^z$. The eigenvalues are:
\begin{eqnarray*}
    \lambda_1 &=& 0.7846   \\
  \lambda_{2,3} &=& -0.028 \pm 0.250 i  \\
  |\lambda_{2,3}| &=&  0.2512.  \\
\end{eqnarray*}
To illustrate the orbit's local exponential stability, the 3D biped's model in
closed-loop is simulated with the initial state perturbed from the fixed-point
$x^\ast $. An initial error of $-1^\circ$ is introduced on each joint and a velocity error of $-5^\circ s^{-1}$ is introduced on each joint velocity.

\begin{figure}[htbp]
\psfrag{q1minus (rad)}{\small{$\qoneonef$ (rad)}}
\psfrag{thminus (rad)}{\small{$\thetaonef$  (rad)}}
\psfrag{dq1minus (rad/sec)}{\small{$\dotqoneonef$  (rad/s)}}
\psfrag{dthminus (rad/sec)}{\small{$\dotthetaonef$ (rad/s)}}
\psfrag{step number}{\small{step number}}
\centerline{\includegraphics[width=3.in]{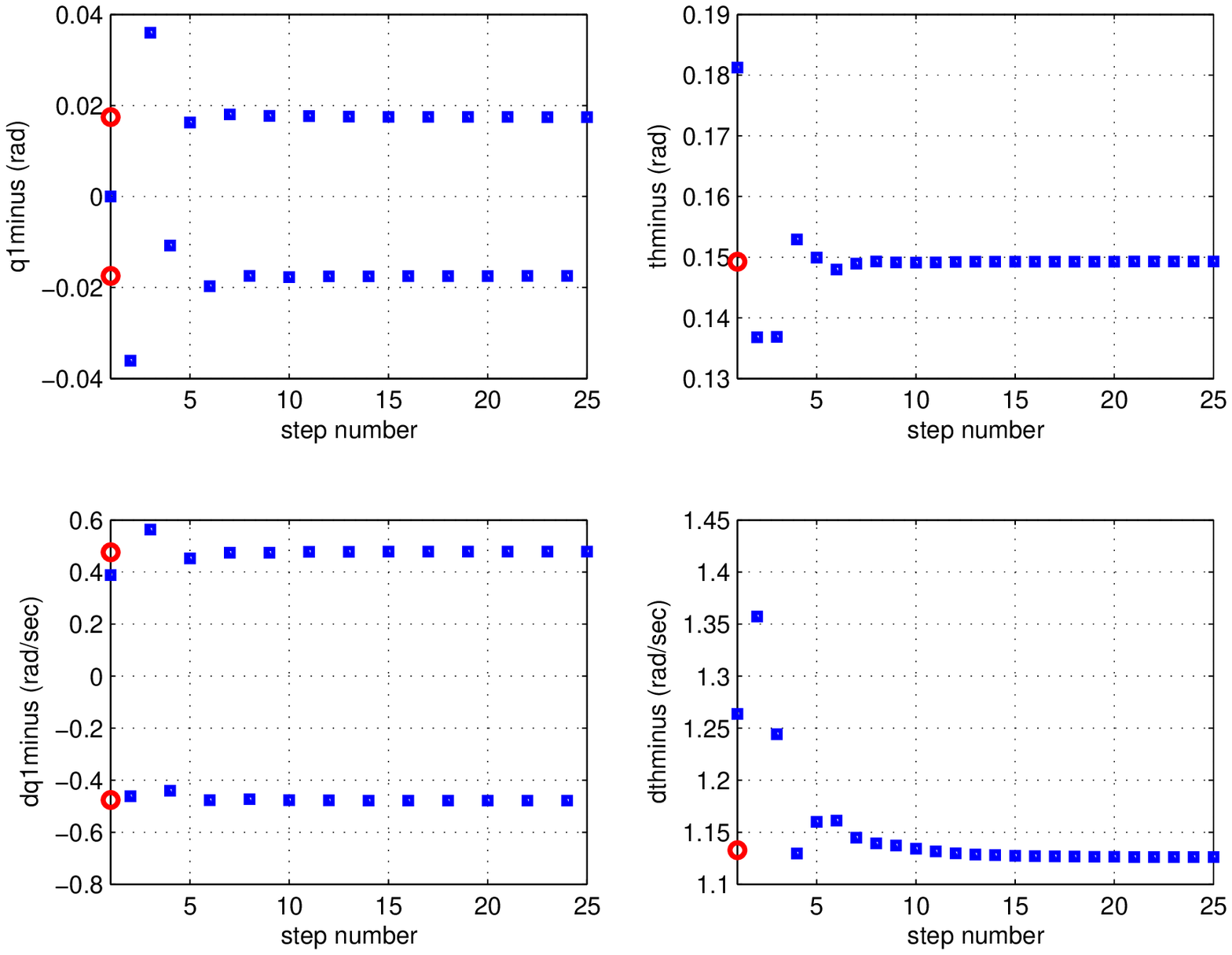}}
\caption{The evolution of $q_u$ at the end of each step for the
3D biped's full model under closed-loop walking control,
with the initial condition perturbed from $x^\ast$. The small circles represent the values on the periodic orbit.}
\label{fig_control_output_qu}
\end{figure}

Fig.~\ref{fig_control_output_qu} shows the
evolution of values of the uncontrolled variables $q_u$ at the end of each step when the new output is used. These variables clearly converge toward the periodic motion.

\begin{figure}[htbp]
\psfrag{q1 (rad)}{\small{$\dot q_1$  (rad/s)}}
\psfrag{qp1 (rad/s)}{\small{$q_1$ (rad)}}
\psfrag{q2 (rad)}{\small{$\dot q_2$  (rad/s)}}
\psfrag{qp2 (rad/s)}{\small{$q_2$ (rad)}}
\psfrag{q3 (rad)}{\small{$\dot q_3$  (rad/s)}}
\psfrag{qp3 (rad/s)}{\small{$q_3$ (rad)}}
\psfrag{q4 (rad)}{\small{$\dot q_4$  (rad/s)}}
\psfrag{qp4 (rad/s)}{\small{$q_4$ (rad)}}\centerline{\includegraphics[width=3.in]{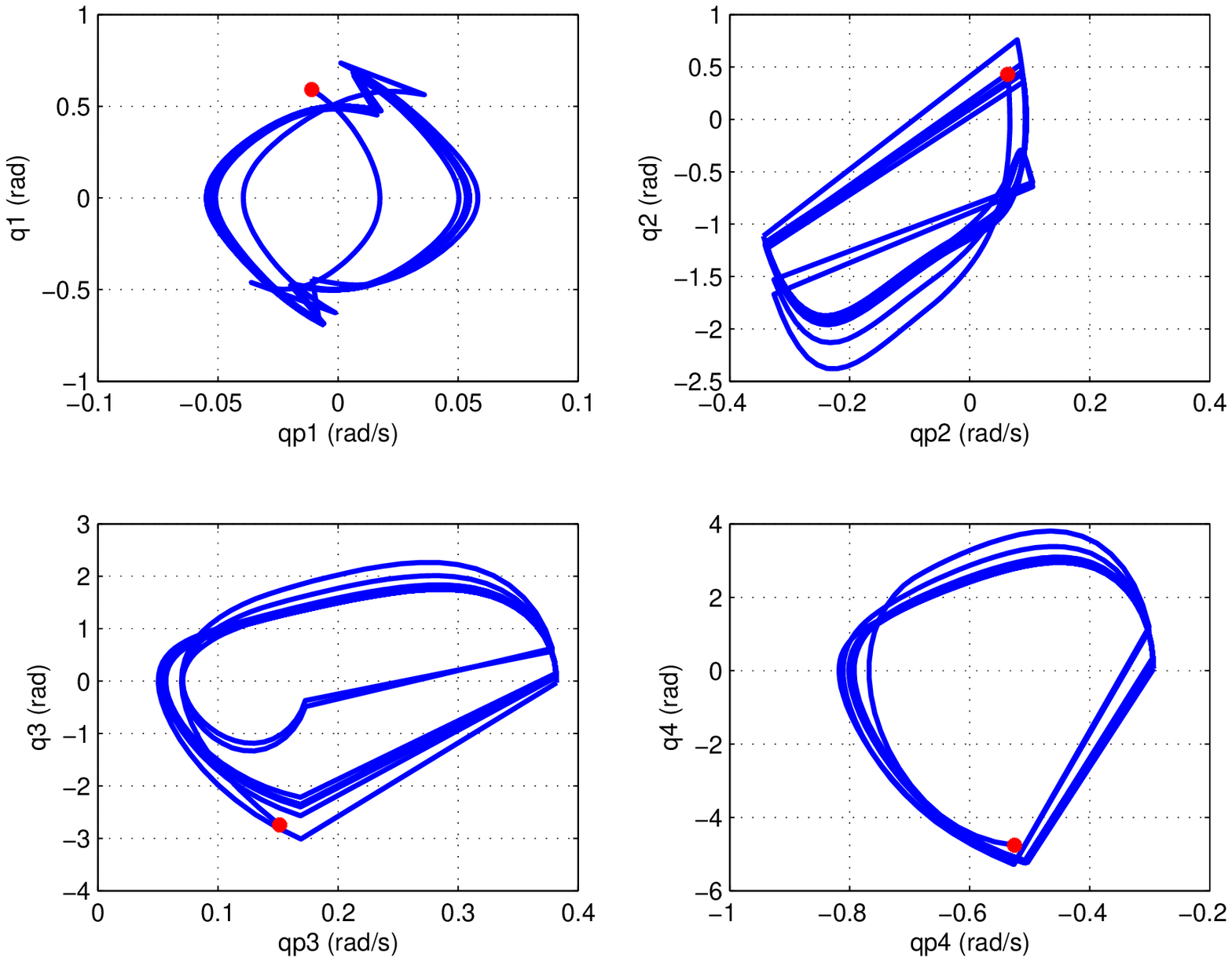}}
\caption{Phase-plane plots for $q_i$, $i=1,\ldots, 4$. The straight lines correspond to the impact phase, where the state of the robot changes instantaneously. The initial state is represented by a (red) star. Each variable converges to a periodic motion.}
\label{fig_control_output_state}
\end{figure}

Fig.~\ref{fig_control_output_state} shows phase-plane plots of the first four variables. The convergence towards a periodic motion is clear for the controlled and uncontrolled variables.

\section{Conclusions}
\label{sec:SGC:conclusion}

A simple 3D bipedal model has been studied, with the objective of
developing a time-invariant feedback control law that induces
asymptotically stable walking, without relying on the use of large
feet. For this reason, a biped consisting of five links, connected to form two legs with knees
and a torso, was assumed to have point feet with
no actuation between the feet and ground. Inspired by its success in
solving similar problems for planar robots, the method of virtual
constraints was applied to the 3D robot, with the virtual
constraints chosen via optimization as suggested in \cite{WEGRKO03}.

The main contributions of the paper are:
\begin{enumerate}
\item The development of an efficient optimization process to obtain periodic motions in 3D for a robot with point contact feet.
\item The computation of human-like periodic walking motions that can be stable or unstable, depending on the choice of actuated variables and corresponding virtual constraints.
\item The numerical study of stability on the basis of a low-dimensional subsystem corresponding to the hybrid zero dynamics. The \Poincare\  return map was computed in a space of dimension three for a robot with two degrees of underactuation.
\item The use of a stride-to-stride controller to stabilize a walking motion that was not naturally stable for a given choice of controlled outputs.
\item The discovery of the importance of the selection of the controlled outputs on the stability of a given periodic motion.
\end{enumerate}

Points (1), (3) and (4) can be viewed as direct extensions of the work previously done on planar walking robots. However, the points (2) and (5) are specific to the study of robots in 3D.

The control strategy developed here for the control
of robots with point feet can be extended to the case of robots with actuated, non-trivial feet, as was done
in  \cite{CHOIJ05a}, \cite{DjChGr07}, \cite{ChDjGr08} for planar robots. In this case, foot
rotation can be included as part of a natural gait and the explicit control of the ZMP position can be addressed. Another possible extension would be to continue with the unactuated point-foot hypothesis, this time including yaw rotation about the foot. It is unclear at this time whether  this extension is straightforward or not.

\vspace{.1cm}
\textsc{Acknowledgment:}
The work of C. Chevallereau is supported by ANR grants for the PHEMA project.
The work of J.W. Grizzle is supported by NSF
grant ECS-0600869. The work of C.L. Shih is supported by Taiwan NSC grant
NSC-96-2221-E-011-126.

\bibliographystyle{IEEEtran}
\bibliography{biped}

\end{document}